\documentclass[sigconf]{aamas} 

\usepackage{balance} 

\usepackage{amsmath,amsfonts,bm}

\def\eqref#1{equation~\ref{#1}}

\def\1{\bm{1}}

\def\vpi{{\bm{\pi}}}

\def\va{{\bm{a}}}

\def\vs{{\bm{s}}}

\DeclareMathAlphabet{\mathsfit}{\encodingdefault}{\sfdefault}{m}{sl}
\SetMathAlphabet{\mathsfit}{bold}{\encodingdefault}{\sfdefault}{bx}{n}

\newcommand{\clip}{\mathrm{clip}}

\newcommand{\TV}{D_{\mathrm{TV}}}
\newcommand{\TVmax}{D_{\mathrm{TV}}^{\mathrm{max}}}

\usepackage{amsmath}
\usepackage{amsthm}
\usepackage{mathtools}
\usepackage{algorithm}
\usepackage{algorithmic}
\usepackage{multirow}
\usepackage{hyperref}
  
\usepackage[capitalize,noabbrev]{cleveref}

\theoremstyle{plain}
\newtheorem{theorem}{Theorem}[section]
\newtheorem{proposition}[theorem]{Proposition}

\theoremstyle{definition}
\newtheorem{definition}[theorem]{Definition}
\newtheorem{assumption}[theorem]{Assumption}
\theoremstyle{remark}
\newtheorem{remark}[theorem]{Remark}

\newcommand{\norm}[1]{\left\lVert#1\right\rVert}
\newcommand{\red}[1]{\textcolor{red}{#1}}
\newcommand{\blue}[1]{\textcolor{blue}{#1}}

\setcopyright{ifaamas}
\acmConference[AAMAS '23]{Proc.\@ of the 22nd International Conference
on Autonomous Agents and Multiagent Systems (AAMAS 2023)}{May 29 -- June 2, 2023}
{London, United Kingdom}{A.~Ricci, W.~Yeoh, N.~Agmon, B.~An (eds.)}
\copyrightyear{2023}
\acmYear{2023}
\acmDOI{}
\acmPrice{}
\acmISBN{}

\acmSubmissionID{587}

\title[Trust Region Bounds for Decentralized PPO]{Trust Region Bounds for Decentralized PPO Under Non-stationarity}

\author{Mingfei Sun}
\affiliation{
  \institution{University of Manchester}
  \city{Manchester}
  \country{United Kingdom}}
\email{mingfei.sun@manchester.ac.uk}
\additionalaffiliation{
  University of Oxford and Microsoft Research when this work was done
}

\author{Sam Devlin}
\affiliation{
  \institution{Microsoft Research}
  \city{Cambridge}
  \country{United Kingdom}}
\email{sam.devlin@microsoft.com}

\author{Jacob Beck}
\affiliation{
  \institution{University of Oxford}
  \city{Oxford}
  \country{United Kingdom}}
\email{jacob.beck@linacre.ox.ac.uk}

\author{Katja Hofmann}
\affiliation{
  \institution{Microsoft Research}
  \city{Cambridge}
  \country{United Kingdom}}
\email{katja.hofmann@microsoft.com}

\author{Shimon Whiteson}
\affiliation{
  \institution{University of Oxford}
  \city{Oxford}
  \country{United Kingdom}}
\email{shimon.whiteson@cs.ox.ac.uk}

\begin{abstract}
We present trust region bounds for optimizing decentralized policies
in cooperative Multi-Agent Reinforcement Learning (MARL), 
which holds even when the transition dynamics are non-stationary. 
This new analysis provides a theoretical understanding of the strong performance of 
two recent actor-critic methods for MARL, which both rely on \emph{independent ratios}, i.e., computing probability ratios separately for each agent's policy. We show that, 
despite the non-stationarity that independent ratios cause,
a monotonic improvement guarantee still arises 
as a result of enforcing the trust region constraint over all decentralized policies. 
We also show this trust region constraint 
can be effectively enforced in a principled way by bounding independent ratios 
based on the number of agents in training, 
providing a theoretical foundation for proximal ratio clipping. 
Finally, our empirical results support the hypothesis 
that the strong performance of IPPO and MAPPO is a direct result of enforcing 
such a trust region constraint via clipping in centralized training, 
and tuning the hyperparameters with regards to the number of agents,
as predicted by our theoretical analysis. 
\end{abstract}


\newcommand{\BibTeX}{\rm B\kern-.05em{\sc i\kern-.025em b}\kern-.08em\TeX}

\begin{document}


\pagestyle{fancy}
\fancyhead{}


\maketitle 


\section{Introduction}
In cooperative multi-agent reinforcement learning (MARL), 
a team of agents must coordinate their behavior to maximize 
a single cumulative return~\cite{panait_cooperative_2005}. 
In such a setting, partial observability and/or communication constraints 
necessitate the learning of decentralized policies that condition only on 
the local action-observation history of each agent.
In a simulated or laboratory setting, decentralized policies can often be learned 
in a centralized fashion, i.e., Centralized Training with Decentralized Execution (CTDE)\cite{oliehoek_concise_2016}, 
which allows agents to access each other's observations and 
unobservable extra state information during training.

Actor-critic algorithms~\cite{konda2000actor} are a natural approach to CTDE 
because critics can exploit centralized training by conditioning on extra information 
not available to the decentralized policies~\cite{lowe2017multi,foerster_counterfactual_2017}. 
Unfortunately, such actor-critic methods have long been outperformed by 
value-based methods such as QMIX~\cite{rashid2018qmix} on MARL benchmark tasks such as Starcraft Multi-Agent Challenge (SMAC)~\cite{samvelyan2019starcraft}. 
However, two recent actor-critic algorithms~\cite{de2020independent,yu2021surprising}
have upended this ranking by outperforming previously dominant MARL methods, 
such as MADDPG~\cite{lowe2017multi} and value-decomposed 
$Q$-learning~\cite{sunehag_value-decomposition_2017,rashid2018qmix}.

Both algorithms are multi-agent extensions of Proximal Policy Optimization 
(PPO)~\cite{schulman2017proximal} but one uses decentralized critics, 
i.e., independent PPO (IPPO)~\cite{de2020independent}, 
and the other uses centralized critics, i.e., multi-agent PPO (MAPPO)~\cite{yu2021surprising}.
One key feature of PPO-based methods is the use of ratios 
(between the policy probabilities before and after updating) in the objective.
Both IPPO and MAPPO extend this feature of PPO to the multi-agent setting 
by computing ratios separately for each agent's policy during training, 
which we call \emph{independent ratios}.
However, until now there has been no theoretical justification for the use of such independent ratios. 

In this paper we show that the analysis that 
underpins the monotonic policy improvement guarantee for PPO~\cite{schulman2015trust} 
does not carry over to the use of independent ratios in IPPO and MAPPO. 
Instead, a direct application of this analysis leads to a joint policy optimization 
and suggests the use of \emph{joint ratios}, i.e., computing ratios between joint policies.The difference is crucial because, 
based on the existing trust region analysis for PPO, 
only a joint ratios approach enjoys a monotonic policy improvement guarantee.  
Moreover, as independent ratios consider only the change in one agent's policy 
and ignore the fact that the other agents' policies also change, 
the transition dynamics underlying these independent ratios are 
non-stationary~\cite{papoudakis2019dealing}, 
breaking the assumptions in the monotonic improvement analysis \cite{schulman2015trust}. 
While some studies attempt to extend the monotonic improvement analysis to 
MARL~\cite{wen2021game,li2020multi}, 
they primarily consider optimizing policies with joint ratios, rather than independent ratios, 
and are thus not applicable to IPPO or MAPPO.

To address this gap, we provide a new monotonic improvement analysis 
that holds even when the transition dynamics are non-stationary. 
We show that, despite this non-stationarity, a monotonic improvement guarantee 
still arises as a result of enforcing the trust region constraint over \emph{all decentralized policies}, 
i.e., a centralized trust region constraint. 
In other words, constraining the update of all decentralized policies in centralized training 
addresses the non-stationarity of learning decentralized policies. 
Our analysis implies that independent ratios can also enjoy the same 
performance guarantee as joint ratios 
if the centralized trust region constraint is properly enforced by bounding independent ratios. 
In this way both IPPO and MAPPO can guarantee monotonic policy improvement. 
We provide a theoretical foundation for proximal ratio clipping 
by showing that centralized trust region can be enforced in a principled way 
by bounding independent ratios based on the number of agents in training. 
Furthermore, we show that the surrogate objectives optimized in IPPO and MAPPO 
are essentially equivalent when their critics converge to a fixed point.

Finally, we provide empirical results that support the hypothesis 
that the strong performance of IPPO and MAPPO is a direct result of enforcing 
such a trust region constraint. Particularly, we show that tuning the hyperparameters for the clipping range
is highly sensitive to the number of agents, as together these
effectively determine the size of the centralized trust region.  
Moreover, we show that IPPO and MAPPO have comparable performance 
on SMAC maps with varied difficulty and numbers of agents. 
This comparable performance also implies that the way of training critics 
could be less crucial in practice than enforcing a trust region constraint.

\section{Background}
\label{sec:background}
\subsection{Dec-MDPs} 
We consider a {\em fully cooperative multi-agent task} 
in which a team of cooperative agents choose sequential actions in a stochastic environment. 
It can be modeled as a {\em decentralized Markov decision process} (Dec-MDP), 
defined by a tuple $\{ \mathcal{N} , \mathcal{S}, \mathcal{A}, p, r, d_0, \gamma \}$, 
where $ \mathcal{N} \triangleq \{1,\dots,N\}$ denotes the set of 
$N$ agents and $\vs \in \mathcal{S}\triangleq\mathcal{S}^1\times\mathcal{S}^2\times...\times\mathcal{S}^N$ 
describes the joint state of the environment. 
The initial state $s^{[0]} \sim d_0$ is drawn from distribution $d_0$, 
and at each time step $t$, all agents $k \in \mathcal{N}$ 
choose simultaneously one action $a_k^{[t]} \in \mathcal{A}^k$, 
yielding a joint action 
$\va^{[t]} \triangleq a^{[t]}_1\times a^{[t]}_2\times...\times a^{[t]}_N\in \mathcal{A}\triangleq \mathcal{A}^1\times\mathcal{A}^2\times...\times\mathcal{A}^N$. 
After executing the joint action $\va^{[t]}$ in state $\vs^{[t]}$, 
the next state $\vs^{[t+1]} \sim p(\vs^{[t]}, \va^{[t]})$ is drawn from 
transition kernel $p$ and a collaborative reward $r^{[t]} = r(\vs^{[t]})$ is returned (for notation simplicity, the reward is defined only on state).
In a Dec-MDP, each agent $k \in \mathcal{N}$ has a local state $s^{[t]}_k \in \mathcal{S}^{k}$, 
and chooses its actions with a decentralized policy 
$a_k^{[t]} \sim \pi_k(\cdot|s_k^{[t]})$ based only on its local state. 
The collaborating team of agents aims to learn a \emph{joint policy}, 
$\bm{\pi}(\va^{[t]}|\vs^{[t]}) \triangleq \prod_{k=1}^N \pi_k(a_k^{[t]}|s_k^{[t]})$, 
that maximizes their expected discounted return, 
$J(\bm{\pi}) \triangleq \mathbb{E}_{(\vs^{[t]}, \va^{[t]})}[\sum_{t=0}^{\infty} \gamma^t r^{[t]}]$, 
where $\gamma \in [0, 1)$ is a discount factor. 


\subsection{Policy Optimization Methods} 
For single-agent RL that is modeled as an infinite-horizon discounted Markov decision process (MDP) 
$\{\mathcal{S}, \mathcal{A}, p, r, d_0, \gamma\}$, 
the policy performance is defined as:
$J(\pi) \triangleq \mathbb{E}_{(s^{[t]}, a^{[t]})}\big[ \sum_{t=0}^{\infty}\gamma^t r(s^{[t]}) \big]$. 
The action-value function $q_\pi$ 
and value function $v_\pi$ are defined as:
$q_{\pi}(s^{[t]}, a^{[t]}) = \mathbb{E}_{s^{[t+1]}\sim p(\cdot|s^{[t]},a^{[t]}),a^{[t+1]}\sim\pi(\cdot|s^{[t+1]})}\sum_{l=0}^{\infty} \gamma^l r(s^{[t+l]})$, 
$v_{\pi}(s^{[t]}) = \mathbb{E}_{a^{[t]}\sim\pi(\cdot|s^{[t]})}\Big[q_{\pi}(s^{[t]}, a^{[t]}) \Big]$. 
Define the advantage function as $A_{\pi}(s, a) \triangleq q_{\pi}(s, a) - v_{\pi}(s)$. 
The following useful identity expresses the expected return of another policy $\tilde{\pi}$ 
in terms of the advantage over $\pi$~\cite{kakade2002approximately}:
\begin{equation}
J(\tilde{\pi}) = J(\pi) + \frac{1}{1-\gamma} \sum_{s}d_{\tilde{\pi}}(s)\sum_{a}\tilde{\pi}(a|s) A_{\pi}(s, a),
\end{equation}
where $d_{\tilde{\pi}}(s)$ is the discounted state distribution induced by $\tilde{\pi}$:
$d_{\tilde{\pi}}(s) \triangleq (1-\gamma) \sum_{t=0}^{\infty} \gamma^t \cdot \text{Probability}\big(S^{[t]}=s|\tilde{\pi}\big)$.
The complex dependency of $d_{\tilde{\pi}}(s)$ on $\tilde{\pi}$ 
makes the righthand side difficult to optimize directly. 
\cite{schulman2015trust} proposed to consider the following surrogate objective
\begin{equation}
L_{\pi}(\tilde{\pi}) \triangleq J(\pi) + \frac{1}{1-\gamma} \sum_{s}d_{\pi}(s)\sum_{a}\tilde{\pi}(a|s)A_{\pi}(s, a), 
\end{equation}
where $d_{\tilde{\pi}}(s)$ is replaced with $d_{\pi}(s)$. 

Define $\TVmax(\pi, \tilde{\pi})\triangleq \max_{s} \TV\big(\pi(\cdot|s), \tilde{\pi}(\cdot|s)\big)$, 
where $\TV$ is the total variation (TV) divergence. 
\begin{theorem}\label{theo:single-agent-trust-region}
(Theorem 1 in \cite{schulman2015trust}) Let $\alpha \triangleq \TVmax(\pi, \tilde{\pi})$. 
Then the following bound holds
\begin{equation}
J(\tilde{\pi}) \geq L_{\pi}(\tilde{\pi}) - \frac{4\xi \gamma}{(1-\gamma)^2} \alpha^2,
\end{equation}
where $\xi=\max_{s,a}\lvert A_{\pi}(s, a) \rvert$. 
\end{theorem}
This theorem forms the foundation of policy optimization methods, 
including Trust Region Policy Optimization (TRPO)~\cite{schulman2015trust} 
and Proximal Policy Optimization (PPO)~\cite{schulman2017proximal}. 
TRPO suggests a robust way to take large update steps by using a constraint, 
rather than a penalty, on the TV divergence,
and considers the following practical optimization problem, 
\begin{equation}\label{equ:trpo-objective}
\max_{\tilde{\pi}} \mathbb{E}_{(s, a)\sim d_{\pi}}\Big[\frac{\tilde{\pi}(a|s)}{\pi(a|s)}A_{\pi}(s, a) \Big], 
\quad\text{s.t.} \quad \TVmax(\pi, \tilde{\pi}) \leq \delta, 
\end{equation}
where $\delta$ specifies a TV threshold. 
This constrained optimization is complicated as it requires 
using conjugate gradient algorithms with a quadratic approximation to the constraint. 
PPO simplifies the above optimization by clipping probability ratios 
$\lambda_{\tilde{\pi}}=\frac{\tilde{\pi}(a|s)}{\pi(a|s)}$
to form a lower bound of $L_{\pi}(\tilde{\pi})$: 
\begin{equation}\label{equ:ppo-objective}
\quad \max_{\tilde{\pi}} \mathbb{E}_{(s, a)\sim d_{\pi}} \big[ \min\big(\lambda_{\tilde{\pi}} A_{\pi}(s, a), 
\clip(\lambda_{\tilde{\pi}}, 1\pm\epsilon)A_{\pi}(s, a)  \big) \big], 
\end{equation}
where $\epsilon$ is a hyperparameter to specify the clipping range. 

\subsection{Independent PPO and Multi-Agent PPO}
Both IPPO~\cite{de2020independent} and MAPPO~\cite{yu2021surprising} 
optimize decentralized policies with independent ratios. 
In particular, assume the policy and the advantage function are parameterized by $\theta$, $\phi$ respectively, 
the main objective IPPO and MAPPO optimize is 
\begin{equation}
\max_{\tilde{\pi}_\theta} \sum_{k} \mathbb{E}_{(s_k, a_k)\sim d_{\pi_\theta}} \big[ \min\big(\lambda_\theta A_\phi(s_k, a_k), 
 \clip(\lambda_\theta, 1\pm\epsilon)A_\phi(s_k, a_k  \big) \big], 
\end{equation}
where $\lambda_\theta=\frac{\tilde{\pi}_\theta(a_k|s_k)}{\pi_\theta(a_k|s_k)}$
denotes the ratio between the decentralized policy probabilities of agent $k$ before and after updating. 
The difference between IPPO and MAPPO lies in how they estimate the advantage function: 
IPPO learns a fully decentralized advantage function 
$A_{\phi}(s_k, a_k) \triangleq \sum_{t=0}^{\infty}[r(s_k^{[t]})] - v_{\pi_k}(s_k)$ 
based on the local information $(s_k, a_k)$ for each agent, 
while MAPPO uses a centralized critic that conditions on centralized state information $\vs$:
$A_{\phi}(s_k, a_k) \triangleq \mathbb{E}_{s_{-k}}\big[\sum_{t=0}^{\infty}[r(s_k^{[t]})] - v_{\vpi}(\vs)\big]$, 
where $-k$ refers the set of all agents except agent $k$. 
Both methods use parameter sharing, and all agents share the same actor and critic networks. 
The use of independent ratios together with parameter sharing has shown strong empirical results 
in various MARL benchmark tasks~\cite{de2020independent,yu2021surprising}.

\section{Trust Region Bounds for MARL}\label{sec:decentralized-guarantee}
\label{sec:centralized-trust-region}
In this section, we first directly apply TRPO's trust region analysis 
to cooperative MARL, which yields joint ratios rather than the independent ratios adopted in IPPO and MAPPO. 
We then show that optimization with independent ratios induces non-stationarity in MARL, 
which breaks the stationarity assumption in the trust region analysis.
Finally, we provide a new analysis that shows how monotonic policy improvement can still 
arise from non-stationary transition dynamics with independent ratios.

\subsection{Optimization with Joint Ratios}
Consider the joint policy $\vpi(\va|\vs)$ and the centralized advantage 
function $A_{\vpi}(\vs, \va) = q_{\vpi}(\vs, \va) - v_{\vpi}(\vs)$. Then, 
the trust region analysis for single-agent RL carries over directly to MARL 
with the surrogate objective as
$L_{\vpi}(\tilde{\vpi}) = J(\vpi) + \frac{1}{1-\gamma} \sum_{\vs} d_{\vpi}(\vs)\sum_{\va}\tilde{\vpi}(\va|\vs) A_{\vpi}(\vs, \va)$. 
One can consider the same optimization for TRPO shown in Equation~\ref{equ:trpo-objective}, 
\begin{equation}\label{equ:jr-trpo}
\max_{\tilde{\vpi}} \mathbb{E}_{(\vs, \va)\sim d_{\vpi}}\Big[\frac{\tilde{\vpi}(\va|\vs)}{\vpi(\va|\vs)}A_{\vpi}(\vs, \va) \Big], 
\quad \text{s.t.} \quad \TVmax(\vpi, \tilde{\vpi}) \leq \delta.
\end{equation} 
The trust region constraint is enforced over joint policies, 
which we refer as a \emph{joint trust region constraint}.
With joint ratios defined as 
$\lambda_{\tilde{\vpi}} = \frac{\tilde{\vpi}(\va|\vs)}{{\vpi}(\va|\vs)} = \prod_{k=1}^{N}\big[ \frac{\tilde{\pi}_k(a_k|s_k)}{\pi_k(a_k|s_k)} \big]$, 
one can simplify the above optimization as PPO to have the following objective, 
\begin{equation}\label{equ:jr-ppo}
\max_{\tilde{\vpi}} \mathbb{E}_{(\vs, \va)\sim d_{\vpi}} \big[ \min\big(\lambda_{\tilde{\vpi}} A_{\vpi}(\vs, \va),
\clip(\lambda_{\tilde{\vpi}}, 1\pm\epsilon)A_{\vpi}(\vs, \va)  \big) \big]. 
\end{equation} 
We call the resulting algorithm Joint Ratio PPO (JR-PPO) (see Appendix Algorithm~\ref{algo:joint-ratio}).
Unlike IPPO and MAPPO, 
JR-PPO consider joint ratios over joint policies,
rather than independent ones. 
This difference is crucial, as joint ratios naturally enjoy the monotonic 
improvement guarantee carried over from the single-agent trust region analysis,
Theorem~\ref{theo:single-agent-trust-region}. 
Furthermore, the objective used in IPPO and MAPPO is not equivalent to the above objective 
as they are lower bounds of different objectives. 
Thus, Theorem~\ref{theo:single-agent-trust-region} does not imply any guarantees for IPPO and MAPPO.

\subsection{Optimization with Independent Ratios}
Optimization with independent ratios, however, induces non-stationarity in MARL. 
When optimizing decentralized policies, 
the environment is non-stationary from the perspective of a single agent since the other agents also
change their policies during training. 
To analyze this non-stationarity, 
we first consider the Markov chain for the local state $s_k$ 
induced by the underlying MDP for agent $k$.
When all agents' policies are updated from $\pi_1,...,\pi_N$ to $\tilde{\pi}_1,...,\tilde{\pi}_N$, 
the state transition distribution of this Markov chain also shifts. 
\begin{definition}[State transition shift]
Define the transition shift from $s_k$ to $s'_k$ for agent $k$ as
\begin{multline}
\Delta^{\tilde{\pi}_1, ..., \tilde{\pi}_N}_{\pi_1, ..., \pi_N}(s'_{k}|s_k) \triangleq \sum_{a_k}\big[ p_{\tilde{\pi}_1,...,\tilde{\pi}_N}(s'_k|s_k, a_k)\tilde{\pi}_k(a_k|s_k) \\
- p_{\pi_1,...,\pi_N}(s'_{k}|s_k, a_k)\pi_k(a_k|s_k)\big], 
\end{multline}
where $p_{\pi_1,...,\pi_N}$ and $p_{\tilde{\pi}_1,...,\tilde{\pi}_N}$ refer to the transition dynamics 
before and after $\pi_k$ is updated.
\end{definition}
In the next subsection, we show that the state transition shift consists of two parts: 
an \emph{exogenous part}, which is caused by the update of other agents' policies
(i.e., the change of transition dynamics from $p_{\pi_k}$ to $p_{\tilde{\pi}_k}$), 
and an \emph{endogenous part}, which is contributed by the update of the agent's own policy 
(i.e., the change of agent $k$'s policy from $\pi_k$ to $\tilde{\pi}_k$). 
The exogenous shift breaks the assumption 
in the monotonic improvement guarantee~\cite{schulman2015trust} that the MDP is stationary. 
Consequently, Theorem~\ref{theo:single-agent-trust-region} no longer 
holds if one optimizes with independent ratios as in IPPO and MAPPO. 
See Appendix~\ref{proof:stationarity-trpo} for detailed analysis.

\subsection{Monotonic Improvement Guarantees for Independent Ratios}
We now provide a new analysis for optimization with independent ratios. 
As the exogenous transition shift breaks
the trust region analysis in TRPO, 
we consider how to handle this exogenous shift in training. 
Specifically, since the exogenous shift is caused by the changes of other agents' policies,
it can  be controlled by constraining the update of 
other agents' policies in centralized training. 
\begin{proposition}\label{prop:transition-shift}
In a Dec-MDP, the transition shift 
$\Delta^{\tilde{\pi}_1, ..., \tilde{\pi}_N}_{\pi_1, ..., \pi_N}(s'_{k}|s_k)$
decomposes as follows:
\begin{multline}
\Delta^{\tilde{\pi}_1, ..., \tilde{\pi}_N}_{\pi_1, ..., \pi_N}(s'_k|s_k) = \Delta^{\tilde{\pi}_1,\pi_2, ..., \pi_N}_{\pi_1,\pi_2, ..., \pi_N}(s'_k|s_k) \\
+ \Delta^{\tilde{\pi}_1, \tilde{\pi}_2,\pi_3,..., \pi_N}_{\tilde{\pi}_1,\pi_2,\pi_3, ..., \pi_N}(s'_k|s_k) +...+ \Delta^{\tilde{\pi}_1, ..., \tilde{\pi}_{N-1},\tilde{\pi}_N}_{\tilde{\pi}_1, ..., \tilde{\pi}_{N-1},\pi_N}(s'_k|s_k).
\end{multline}
\end{proposition}
The proof is given in Appendix~\ref{proof:transition-shift}. 
This proposition implies that the state transition shift at local observation $s_k$ is caused 
by the shifts arising from all decentralized policies. 
This decomposition inspires the derivation of a new monotonic improvement guarantee for 
decentralized policy optimization by enforcing the trust region over all decentralized policies. 
Before presenting the guarantee, we first introduce the definition of useful functions and objectives.
\begin{definition}[Decentralized advantages]
For agent $k$, we define a set of decentralized advantage functions as follows:
\begin{multline}\label{equ:advantage-definition}
A_{\pi_k}^{\pi_j}(s_k, a_k) \triangleq r(s_k) + \\ 
\gamma\sum_{s^\prime_k} p_{\tilde{\pi}_1,...,\tilde{\pi}_{j-1},\pi_j,..., \pi_N}(s^\prime_k|s_k, a_k) v_{\pi_k}(s^\prime_k) 
- v_{\pi_k}(s_k), 
\end{multline}
where $v_{\pi_k}$ is the value function under $\pi_k$ with a stationary MDP. 
\end{definition}
This set of advantage functions is defined differently from the canonical ones in that 
it accounts the nuances in transition models, which is important for deriving the improvement guarantees. 

\begin{definition}[Decentralized Surrogate Objectives]
Define the surrogate objective for decentralized policy $\pi_{k}$ as 
\begin{equation}
U_{\pi_k}(\tilde{\pi}_j) \triangleq \mathbb{E}_{(s_k, a_k)\sim d_{\pi_k}\circ\pi_j }\big[\frac{\tilde{\pi}_j(a_k|s_k)}{\pi_j(a_k|s_k)} - 1  \big] A_{\pi_k}^{\pi_j}(s_k, a_k), 
\end{equation}
where $A_{\pi_k}^{\pi_j}(s_k, a_k)$ is agent-$j$'s advantage defined in~Equation~\ref{equ:advantage-definition}, 
$d_{\pi_k}(s_k)$ is state distribution for $s_k$ under $\pi_k$, 
and $(s_k, a_k)\sim d_{\pi_k}\circ\pi_j$ refers to $s_k\sim d_{\pi_k}(s_k), a_k\sim\pi_j(\cdot|s_k)$. 
\end{definition}

\begin{definition}[Objective for Decentralized Policies]
Define the expected return of decentralized policy $\pi_k$ as
\begin{equation}
J(\pi_k) \triangleq \mathbb{E}_{s^{[0]}_k\sim p_0(s_k)}\big[ v_{\pi_k}(s^{[0]}_k) \big], 
\end{equation}
where $p_0(s_k)$ refers to the distribution of starting state $s_k^{[0]}$. 
\end{definition}
\begin{assumption}
The advantage function defined over $(s_k, a_k)$ is bounded under any transition model, 
i.e., $\left|A_{*}^{*}(s_k, a_k)\right|\leq \xi$ for $\forall k\in \mathcal{N}$, 
where $*$ refers to any transition model and decentralized policies considered in the above advantage definition. 
\end{assumption}

We now bound the performance difference between $\tilde{\pi}_k$ and $\pi_k$
with a centralized trust region constraint. 
\begin{theorem}\label{theo:monotonic-decentralized}
Let $\alpha \triangleq  \sum_{j=1}^{N} \mathbb{E}_{s_k\sim d_{\pi_k}}[\TV\big(\pi_j(\cdot|s_k), \tilde{\pi}_j(\cdot|s_k) \big)]$. 
Then the following bound holds for $\forall k \in\mathcal{N}$:
\begin{equation}
J(\tilde{\pi}_k) - J(\pi_k) \geq \frac{1}{1-\gamma}\bigg\{ \sum^N_{j=1} U_{\pi_k}(\tilde{\pi}_j) - \frac{2N\gamma\xi\alpha}{1-\gamma} \bigg\}.
\end{equation}
\end{theorem}
The proof is given in the appendix~\ref{proof:monotonic-decentralized}. 
This theorem implies that, for sufficiently small $\alpha$, 
the performance increase of a decentralized policy $\pi_k$ is lower bounded by 
the sum of surrogate objectives for each decentralized policy with respect to 
the samples generated by $\pi_k$. 
In other words, if the trust region is enforced, 
the sum of surrogate objectives yields an approximate lower bound for 
$J(\tilde{\pi}_k)$, which holds for any decentralized policy $\tilde{\pi}_k$.

Theorem~\ref{theo:monotonic-decentralized} differs from Theorem~\ref{theo:single-agent-trust-region} 
in three respects. 
First, the lower bound for one decentralized policy effectively relies on surrogate objectives for all agents, since the update of one agent's policy 
affects all other agents' transition probability. 
Therefore, to improve the performance for policy $\pi_k$, 
we can simultaneously maximize 
$U_{\pi_k}(\tilde{\pi}_1) + U_{\pi_k}(\tilde{\pi}_2) + ... + U_{\pi_k}(\tilde{\pi}_N)$
on state distribution induced by $\pi_k$. 
Second, unlike the surrogate objective in Theorem~\ref{theo:single-agent-trust-region},
the new surrogate objective explicitly contains an independent ratio 
$\lambda_{\tilde{\pi}_j}\triangleq\frac{\tilde{\pi}_j(a_k|s_k)}{\pi_j(a_k|s_k)}$ 
as it can be rewritten as follows:
$U_{\pi_k}(\tilde{\pi}_{j}) = \mathbb{E}_{(s_k, a_k)}\Big[\big(\frac{\tilde{\pi}_{j}(a_k|s_k)}{\pi_j(a_k|s_k)} - 1\big) A_{\pi_k}^{\pi_j}(s_k, a_k)\Big]$. 
Third, the additional term $\frac{2N\gamma\xi\alpha}{1-\gamma} $ requires 
computing the total variation across all decentralized policies:
$\TV\big(\pi_i(\cdot|s_k), \tilde{\pi}_i(\cdot|s_k) \big)$ for $\forall i\in\mathcal{N}$,
rather than the policies that are directly optimized. 
We call this \emph{centralized trust region}, 
and show in the next section that, in centralized training, this requirement is easily satisfied. 

It is worth noting that, according to the definition of $A_{\pi_k}^{\pi_j}(s_k, a_k)$ in Equation~\ref{equ:advantage-definition}, 
samples $(s_k, a_k)$ should be from the distribution with transient agent-specific transition models, 
which is however infeasible for practical MARL algorithms. 
Instead, we can use the same set of samples from one transition model to obtain a biased estimate of $A_{\pi_k}^{\pi_j}(s_k, a_k)$, 
In the following sections, we show that the parameter sharing can be used to derive practical algorithms. 

\subsection{Trust Regions via Ratio Bounding}\label{sec:ratio-guarantee}
Theorem~\ref{theo:monotonic-decentralized} indicates that 
the centralized trust region is crucial to guarantee monotonic improvement. 
In this section, we show that bounding independent ratios
is an effective way to enforce such centralized trust region constraint, 
and this enforcement requires taking into account the number of agents. 
To achieve this, we first present one proposition about $\TV$ divergence. 

\begin{proposition}\label{theo:trust-region-clipping}
If independent ratios 
$\lambda_{\pi_j} \triangleq \frac{\tilde{\pi}_j(a_k|s_k)}{\pi_j(a_k|s_k)}$
are within the range $[\frac{1}{1+\epsilon_j} , 1+\epsilon_j]$ for $\forall j\in\mathcal{N}$,
then the following bound holds:
\begin{equation}
\mathbb{E}_{s_k\sim d_{\pi_k}}\big[\TV(\pi_j(\cdot|s_k), \tilde{\pi}_j(\cdot|s_k)) \big] \leq \epsilon_j. 
\end{equation}
\end{proposition}

This proposition comes from a property of $\TV$ divergence:
$\TV(\mu(x), \nu(x))=\sum_{\mu(x)>\nu(x)}[\mu(x)-\nu(x)]$ 
where $\mu$ and $\nu$ are two distributions. 
The proof is given in Appendix~\ref{proof:trust-region-clipping}. 
Proposition~\ref{theo:trust-region-clipping} implies that 
bounding independent ratios 
$\frac{\tilde{\pi}_j(a_k|s_k)}{\pi_j(a_k|s_k)}$ with $[\frac{1}{1+\epsilon_j}, 1+\epsilon_j]$ 
amounts to enforcing a trust region constraint with size $\epsilon_j$ over decentralized policies. 
In centralized training, one way to enforce trust region constraint is to delegate 
the centralized trust region constraint to each agent, 
such that the update of each policy $\pi_k(a_k|s_k)$ is bounded. 
One can impose a sufficient condition as follows,
$\mathbb{E}_{s\sim d_{\pi_k}}[\TV\big(\pi_j(\cdot|s_k), \tilde{\pi}_j(\cdot|s_k) \big)]\leq \frac{\delta}{N}$. 
Specifically, with the parameter sharing technique~\cite{gupta2017cooperative}, 
we can update all agents' policies simultaneously 
with the experience from all agents,
\begin{align}
\max_{\theta} &\quad \sum_{k} \mathbb{E}_{(s_k, a_k)\sim d_{\pi_\theta}} \big[\frac{\tilde{\pi}_\theta(a_k|s_k)}{\pi_\theta(a_k|s_k)}  - 1 \big] A_{\phi}(s_k, a_k), \label{equ:parameter-sharing-objective}\\
\text{s.t.} &\quad \mathbb{E}_{s\sim d_{\pi_\theta}}[\TV\big(\pi_\theta(\cdot|s_k), \tilde{\pi}_\theta(\cdot|s_k) \big)] \leq \frac{\delta}{N} , \label{equ:centralized-trust-region-constraint}
\end{align}
where $\theta$ and $\phi$ are shared parameters for policies and critics. 

Furthermore, clipping is one of many ways to approximately achieve this sufficient condition, 
with properly tuned clipping range the number of epochs. 
Consequently, 
we can clip the probability ratios of each decentralized policies to 
form a lower bound of the objective in Equation~\ref{equ:parameter-sharing-objective}, 
similar to PPO~\cite{schulman2017proximal}.
With independent ratios $\lambda_k \triangleq \frac{\tilde{\pi}_\theta(a_k|s_k)}{\pi_{\theta}(a_k|s_k)}$, 
we can optimize the following objective:
\begin{equation}
\max_{\theta} \quad \sum_{k} \mathbb{E}_{(s_k, a_k)\sim d_{\pi_\theta}} \big[ \min\big((\lambda_k -1) A_\theta, 
 \text{clip}(\lambda_k-1,  \pm\epsilon)A_\phi \big) \big], 
\end{equation}
which is exactly the objective used by IPPO and MAPPO.

\begin{remark}
Independently (and evenly) clipping ratios is a sufficient (not necessary) condition to enforce the centralized trust region. 
There are various other ways to achieve this. 
For example, we can factorize the centralized trust region according to a certain coordination graph, 
yielding a coordinated trust region algorithm. 
We can also learn to decompose the centralized trust region such that the sample complexity could be further reduced. 
\end{remark}

\subsection{Learning Advantage Functions}
We now look at the training of the advantage function, where IPPO and MAPPO differ. 
IPPO trains a decentralized advantage function, 
while MAPPO trains a centralized one that incorporates centralized state information. 
Assume a stationary distribution of $(s_k, a_k)$ exists. 
From~\cite{lyu2021contrasting}, we have the following:
\begin{proposition}
(Lemma 1 \& 2 in \cite{lyu2021contrasting}) 
Training of centralized critic and $k$-th decentralized critic
admits unique fixed points 
$q_\pi(s_k, s_{-k}, a_k, a_{-k})$ and 
$\mathbb{E}_{s_{-k},a_{-k}}[q_\pi(s_k, s_{-k}, a_k, a_{-k})]$ respectively,
where $q_\pi$ is the true expected return under the joint policy $\pi$. 
\end{proposition}
Accordingly, based on the definition,
the centralized value function is 
$v(\vs)=v(s_k, s_{-k})=\mathbb{E}_{a_k, a_{-k}}[q_\pi(s_k, s_{-k}, a_k, a_{-k})]$ 
and the decentralized one is 
$v(s_k) = \mathbb{E}_{s_{-k},a_k, a_{-k}}[q_\pi(s_k, s_{-k}, a_k, a_{-k})] = \mathbb{E}_{s_{-k}}[v(s_k, s_{-k})] = \mathbb{E}_{s_{-k}}[v(s)]$. 

Thus, we have $A^{\text{IPPO}}(s_k, a_k) = A^{\text{MAPPO}}(s_k, a_k)$ (and so IPPO and MAPPO objectives are equivalent 
given that the underlying critics converge to a fixed point.

\section{Experiments}
\begin{figure}
  \centering
  \includegraphics[width=0.9\linewidth]{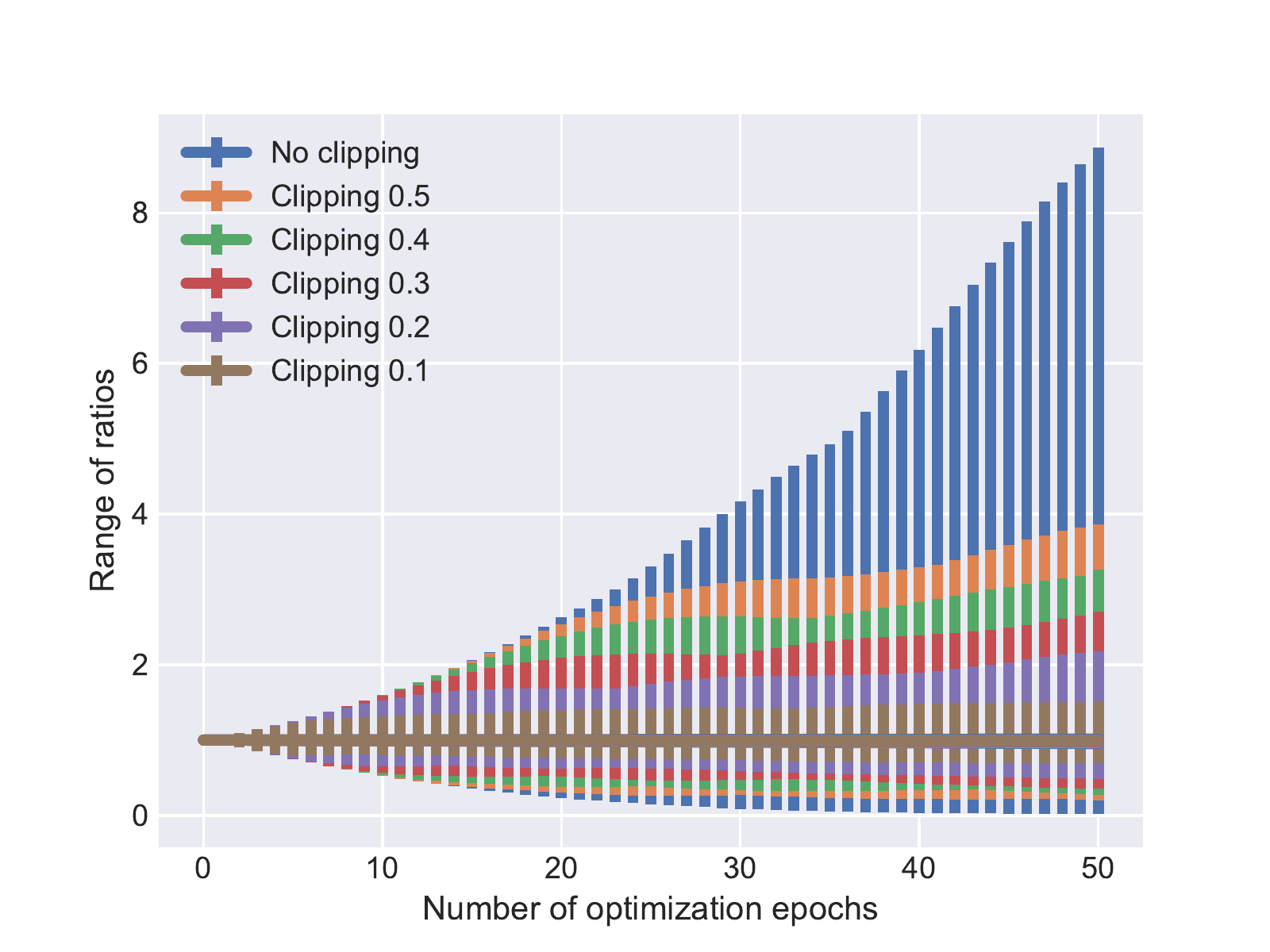}
  \caption{Ratio ranges for 5 agents with the number of optimization epochs;}
  \label{fig:ratio_range_with_epochs}
\end{figure}

We consider the StarCraft Multi-Agent Challenge (SMAC)~\cite{samvelyan2019starcraft} 
for our empirical analysis as it provides a wide range of multi-agent tasks 
with varied difficulty and numbers of agents, see Table~\ref{tab:number-of-agents} for map details. 
We first show that clipping is an effective way to constraint ratios 
when the number of optimization epochs and the learning rate are properly specified. 
Furthermore, we show that clipping also requires taking into account the number of agents 
such that the centralized trust region can be properly enforced. 
We then empirically demonstrate that bounding independent ratios in effect enforces the
trust region over joint policies. 
Finally, we present results showing that IPPO and MAPPO perform equivalently on SMAC maps with varied difficulty and numbers of agents. 

\begin{figure}
  \centering
  \includegraphics[width=0.48\linewidth]{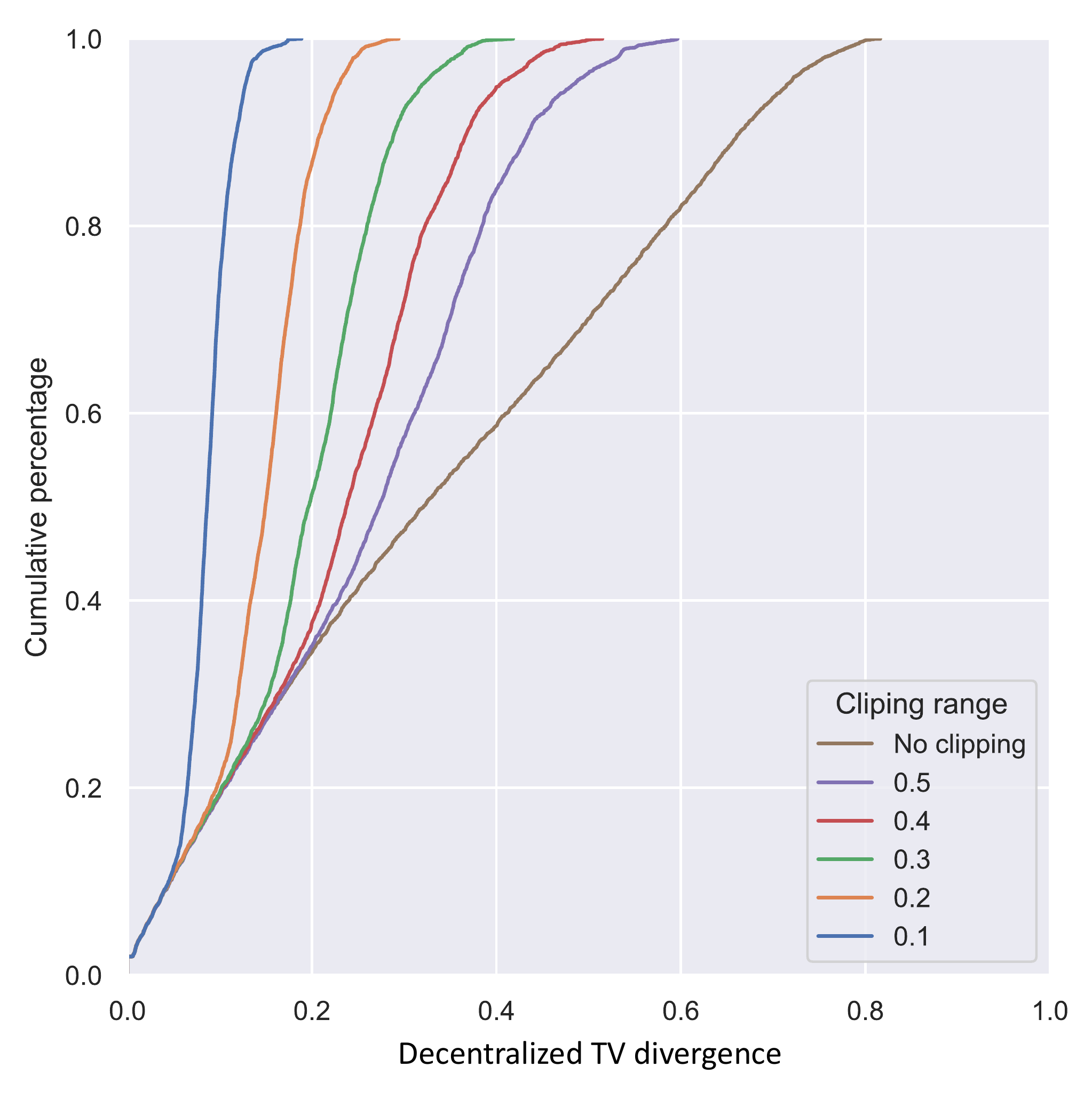}
  \includegraphics[width=0.48\linewidth]{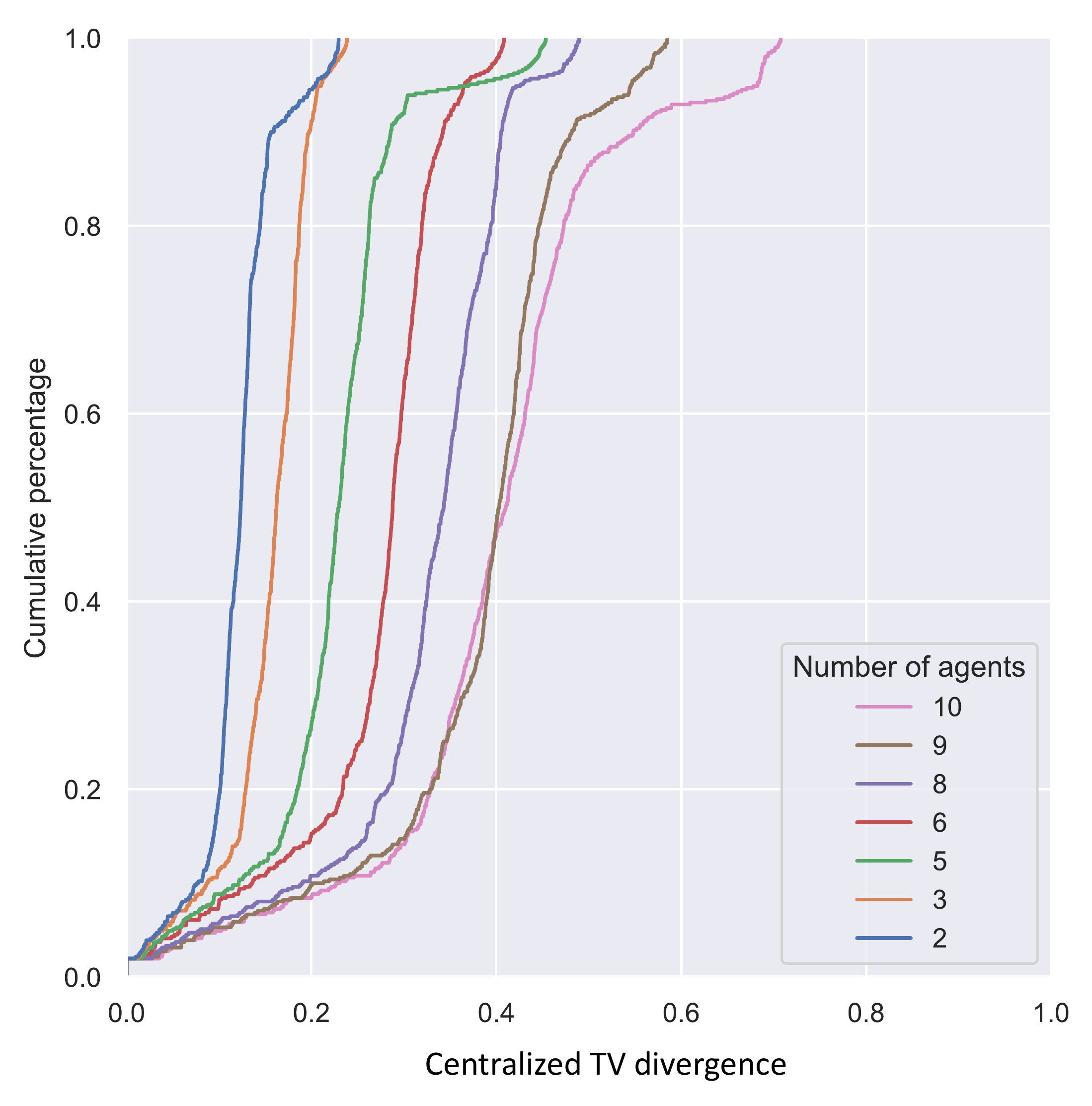}
  \caption{Cumulative percentage of decentralized TV divergence as the clipping value varies (left), 
  and cumulative percentage of centralized TV divergence as the number of agents varies but with fixed clipping range $0.1$ (right).}
  \label{fig:trust_region}
\end{figure}

\subsection{Clipping and Ratio Ranges} 
Proposition~\ref{theo:trust-region-clipping} indicates that bounding independent ratios
amounts to enforcing a trust region constraint over joint policies. 
We empirically show that independent ratio clipping approximately 
bounds independent ratios in the training if the hyperparameters are properly set. 
We train decentralized policies on one map, \emph{2s3z}, and 
clip the independent ratios in the surrogate objective. 
Figure~\ref{fig:ratio_range_with_epochs} shows how the max and min of the ratios 
changes according to the number of optimization epochs with different clipping values. 
Independent ratio clipping can effectively 
constrain the range of ratios only when the number of optimization epochs 
and the clipping range are properly specified. 
In particular, the range of independent ratios grows 
as the number of optimization epochs increases. 
This growth is slower when the clipping range is smaller, e.g., $\epsilon=0.1$. 
Furthermore, the clipping range may not strictly bound ratios 
between $[\frac{1}{1+\epsilon}, 1+\epsilon]$: 
when the clipping range is $0.1$, the independent ratios can exceed $1.2$; 
and the independent ratios can even grow up to $1.6$ when the clipping range is $0.3$. 
We also present the more results on small clipping values in Appendix~\ref{app:small-clipping-range}. 
It is true that a small clipping value results in a small trust region, 
However, when the clip value is too small, 
the resulting trust region makes the update step in each iteration 
also too small to improve the policy. 
Thus, one would need to trade off between the trust region constraint, 
to ensure monotonic improvement, and the policy update step, 
to ensure a sufficient parameter change at each iteration. 

\begin{figure*}
  \centering
  \includegraphics[width=0.80\linewidth]{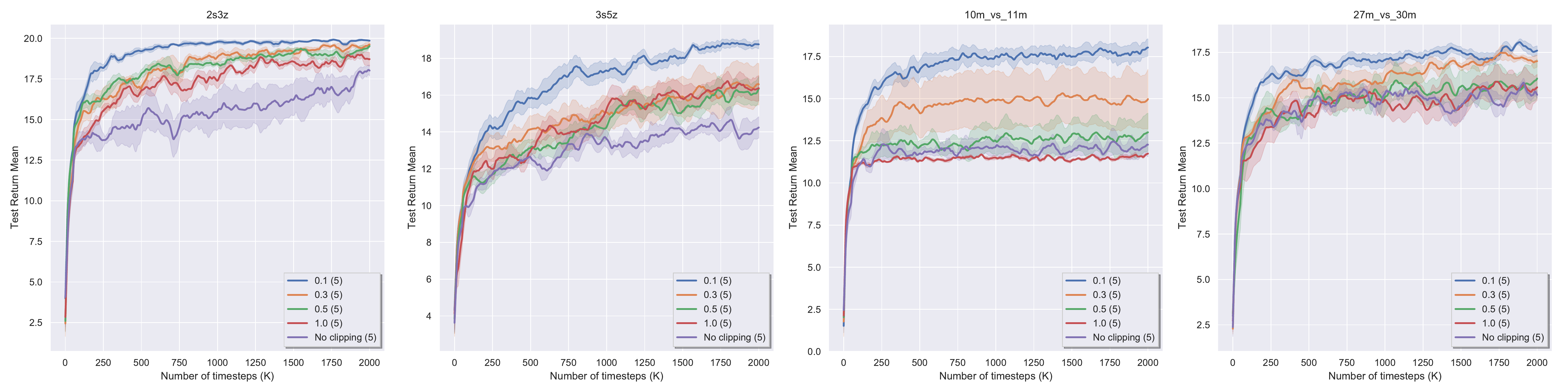}
  \includegraphics[width=0.80\linewidth]{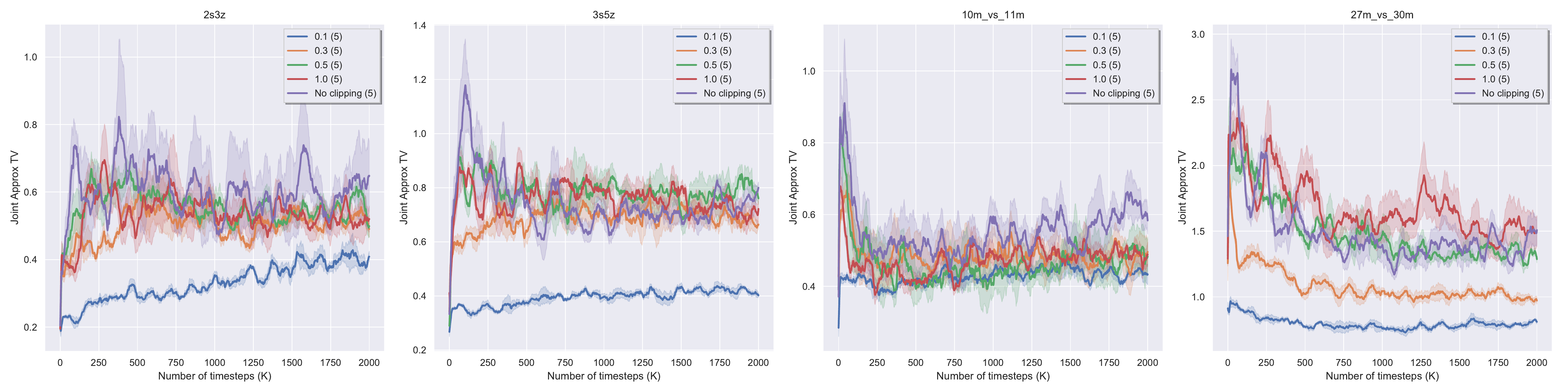}
  \caption{Empirical returns and trust region estimates for independent ratio clipping.}
  \label{fig:clipping_ablation_on_smac}
\end{figure*}

\subsection{Ratio Clipping and Trust Region Constraint} 
Next, we show that the trust region defined by the total variation is empirically 
bounded by independent ratio clipping, 
and this bound is proportional to the number of agents. 
Specifically, we compute the average total variation divergence $\TV$ 
over empirical samples collected by the behavior policy during the first round of actor update,
which contains 100 optimization epochs, 
and report the distribution of $\TV$. 
Figure~\ref{fig:trust_region}(left) shows the distribution 
of $\TV$ over decentralized policies when clipping range varies. 
For clipping at $0.1$, all average $\TV$ values are smaller than $0.2$, 
meaning that the trust region is effectively enforced to be small. 
As the clipping range increases, more $\TV$ values exceed $0.3$. 
For the case without clipping, $\TV$ almost uniformly distributes among $[0.0, 0.8]$, 
implying trust region is no longer enforced. 
Figure~\ref{fig:trust_region}(right) presents the distribution of centralized $\TV$ over all decentralized polices
for clipping at $0.1$, on maps with different number of agents. 
See appendix Table~\ref{tab:number-of-agents} for more details on agent numbers. 
The $\sum_{i=1}^{N}\big(\TV(\pi_i, \tilde{\pi}_i))$ is estimated by summing up 
the empirical total variation distances $\TV(\pi_k, \tilde{\pi}_k)$ over all agents. 
The $\sum_{i=1}^{N}\big(\TV(\pi_i, \tilde{\pi}_i))$ grows almost proportionally 
with the number of agents, 
indicating that enforcing the centralized trust region with independent ratio clipping 
also requires considering the number of agents. 
Figure~\ref{fig:joint_trust_region_epoch} in Appendix~\ref{appen:more-exps} presents the distribution of centralized $\TV$ over all decentralized polices
with different numbers of epochs for clipping at $0.1$. 
Compared to the number of agents, the number of epochs has less impact on the trust region. 
However, as the policy optimization proceeds, 
the impact of the number of epochs on the trust region may increase. 
One may need to tune the learning rate to combat this side-effect~\cite{schulman2017proximal,sun2022you}. 

\begin{figure*}
  \centering
  \includegraphics[width=0.80\linewidth]{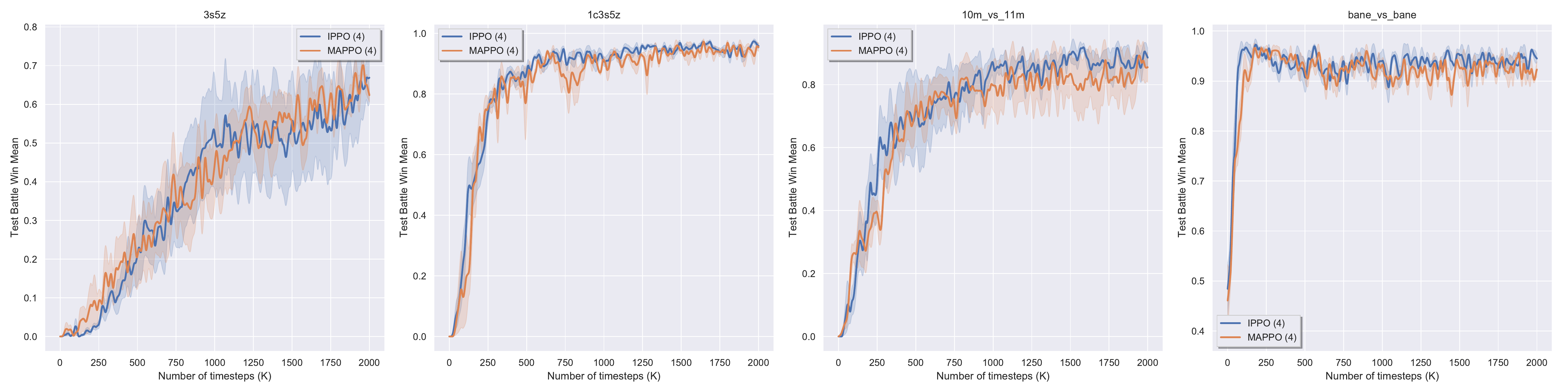}
  \caption{Contrasting IPPO and MAPPO across different maps.}
  \label{fig:ippo-mappo-smac-performance}
\end{figure*}

\subsection{Independent Ratio Clipping on SMAC}
Figure~\ref{fig:clipping_ablation_on_smac} shows the empirical returns and trust region estimates
with different ratio clipping values across different maps in SMAC.
We adopted recurrent networks, i.e., LSTM, as the decentralized policy architecture 
to overcome any partial observability issue in SMAC. 
\footnote{Trained via decentralized advantage, i.e., IPPO.
Results with centralized advantage are similar, 
as presented in Appendix~\ref{appen:more-exps}.
Unlike~\cite{yu2021surprising}, the value function is not clipped. } 
Notably, when the clipping value is small, e.g., $\epsilon=0.1$, 
the joint total variation distance, i.e., the centralized trust region, 
can be effectively bounded, as in the second row in Figure~\ref{fig:clipping_ablation_on_smac}. 
The empirical returns corresponding to $\epsilon=0.1$ are thus improved monotonically, 
outperforming all other returns consistently in four maps.
Moreover, as the number of agents increases,
the trust region enforced by clipping value $\epsilon=0.1$ in the initial training phase 
also grows from less than 0.3 to more than 0.5. 
In contrast, for clipping at $0.5$ and $1.0$, the learning quickly plateaus at local optima, 
especially on maps with many agents, e.g., 10m\_vs\_11m and 27m\_vs\_30m, 
which shows that the policy performance $J(\pi_k)$ is closely related to 
the enforcement of trust region. 
In addition, the test battle win mean of IPPO is presented in Figure~\ref{fig:ippo-test-battle-win-mean} in Appendix~\ref{appen:more-exps}.

\begin{figure*}
   \centering
   \includegraphics[width=0.40\linewidth]{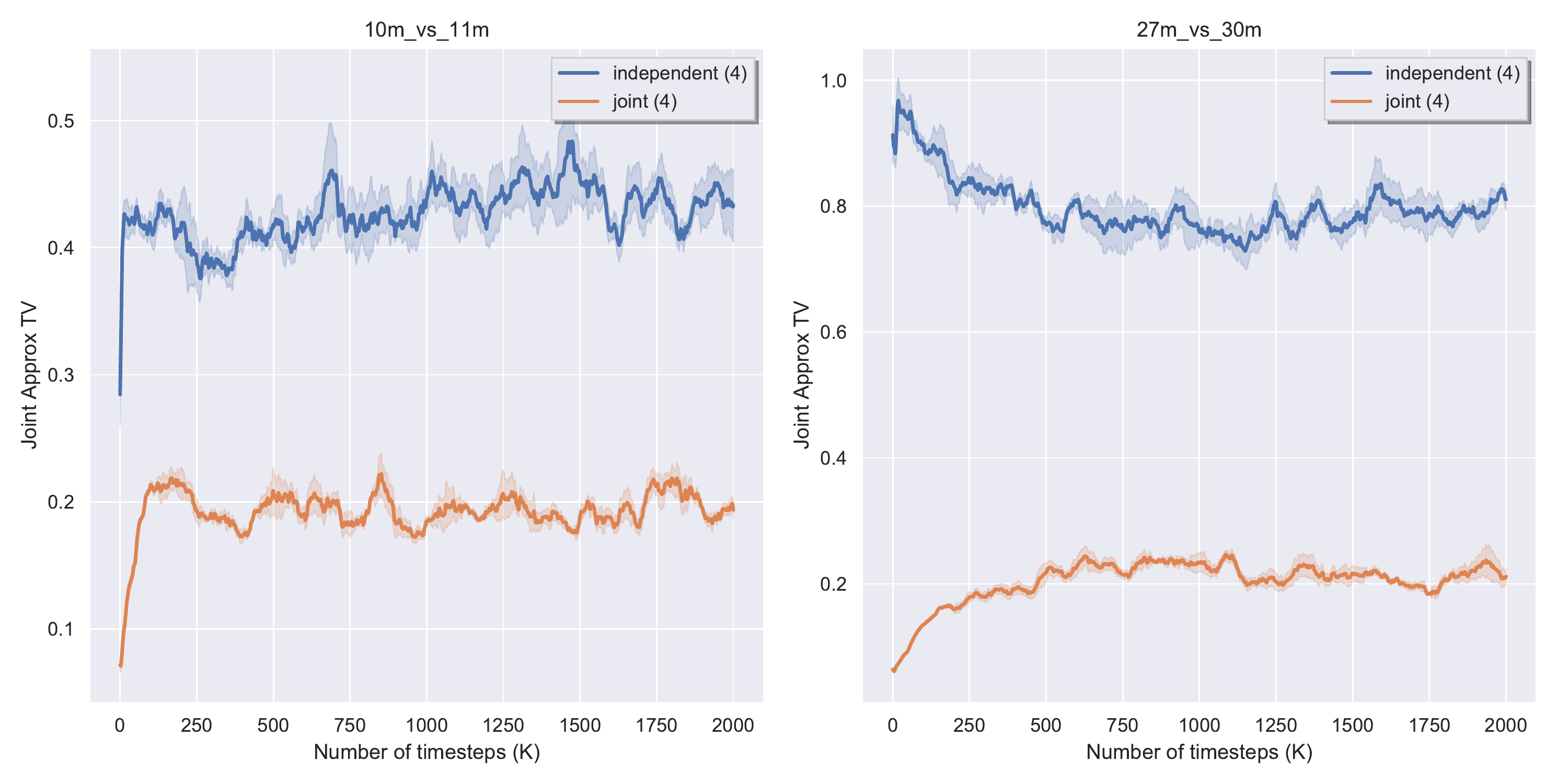}
   \includegraphics[width=0.40\linewidth]{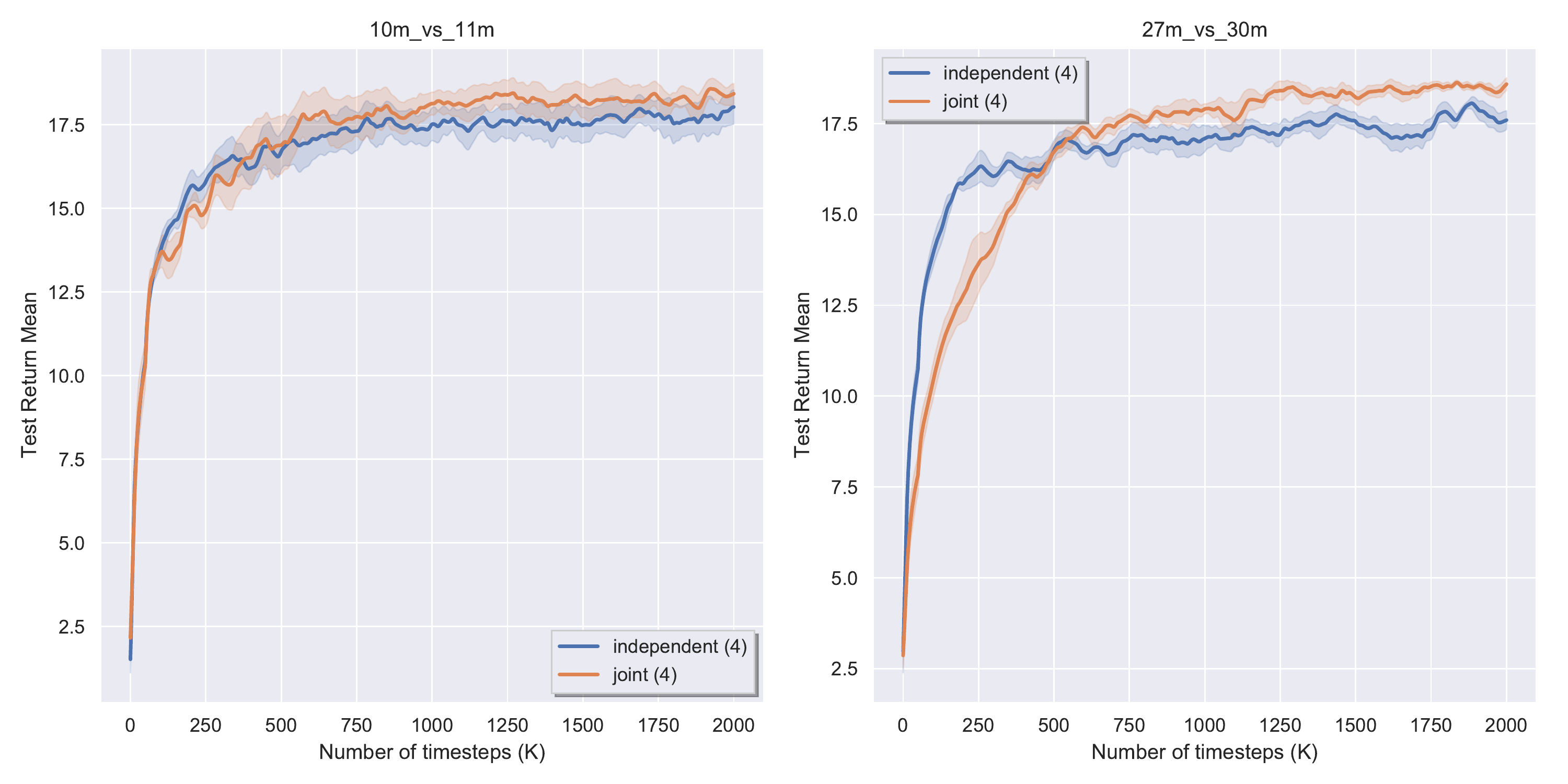}
   \includegraphics[width=0.40\linewidth]{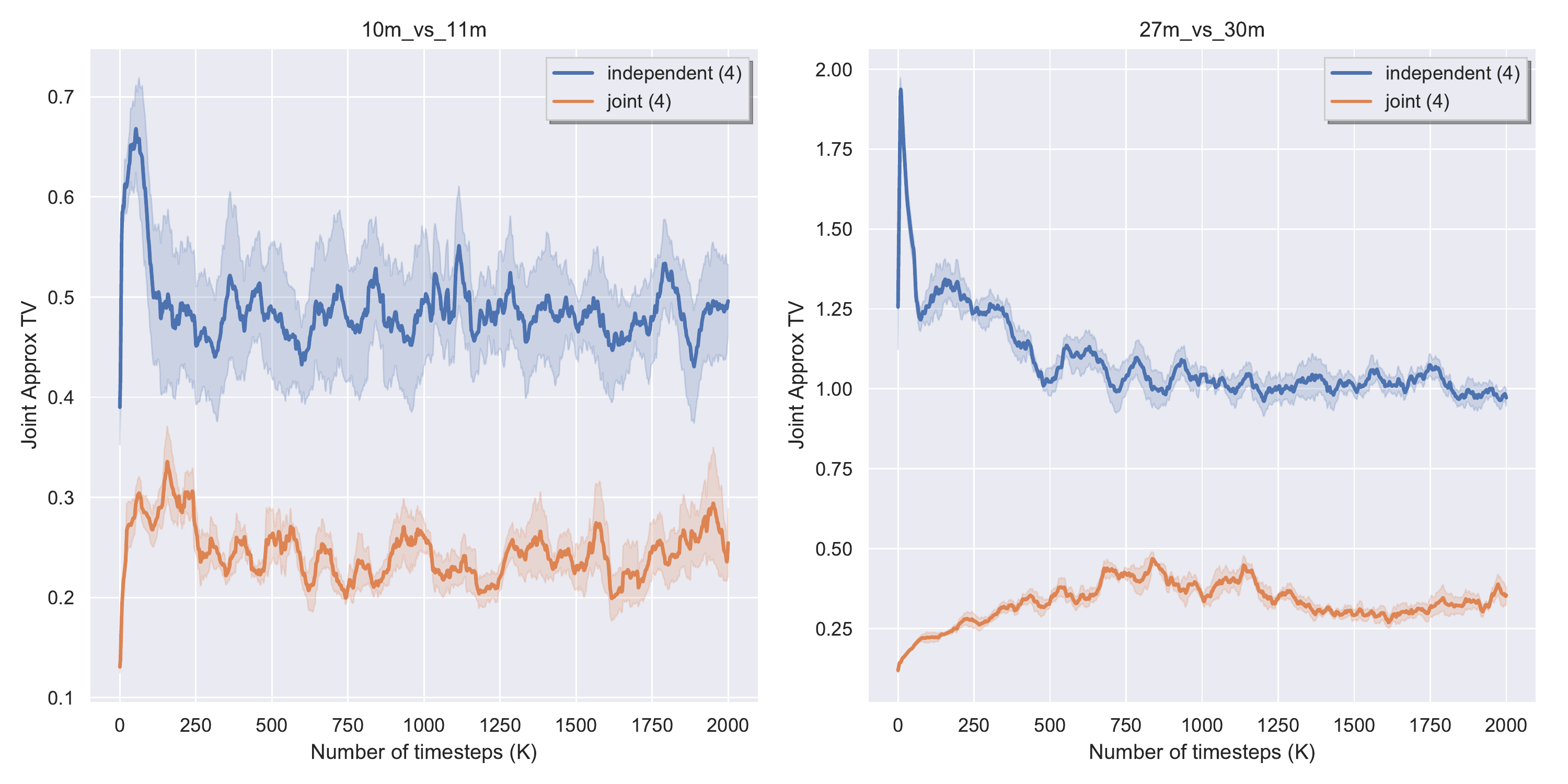}
   \includegraphics[width=0.40\linewidth]{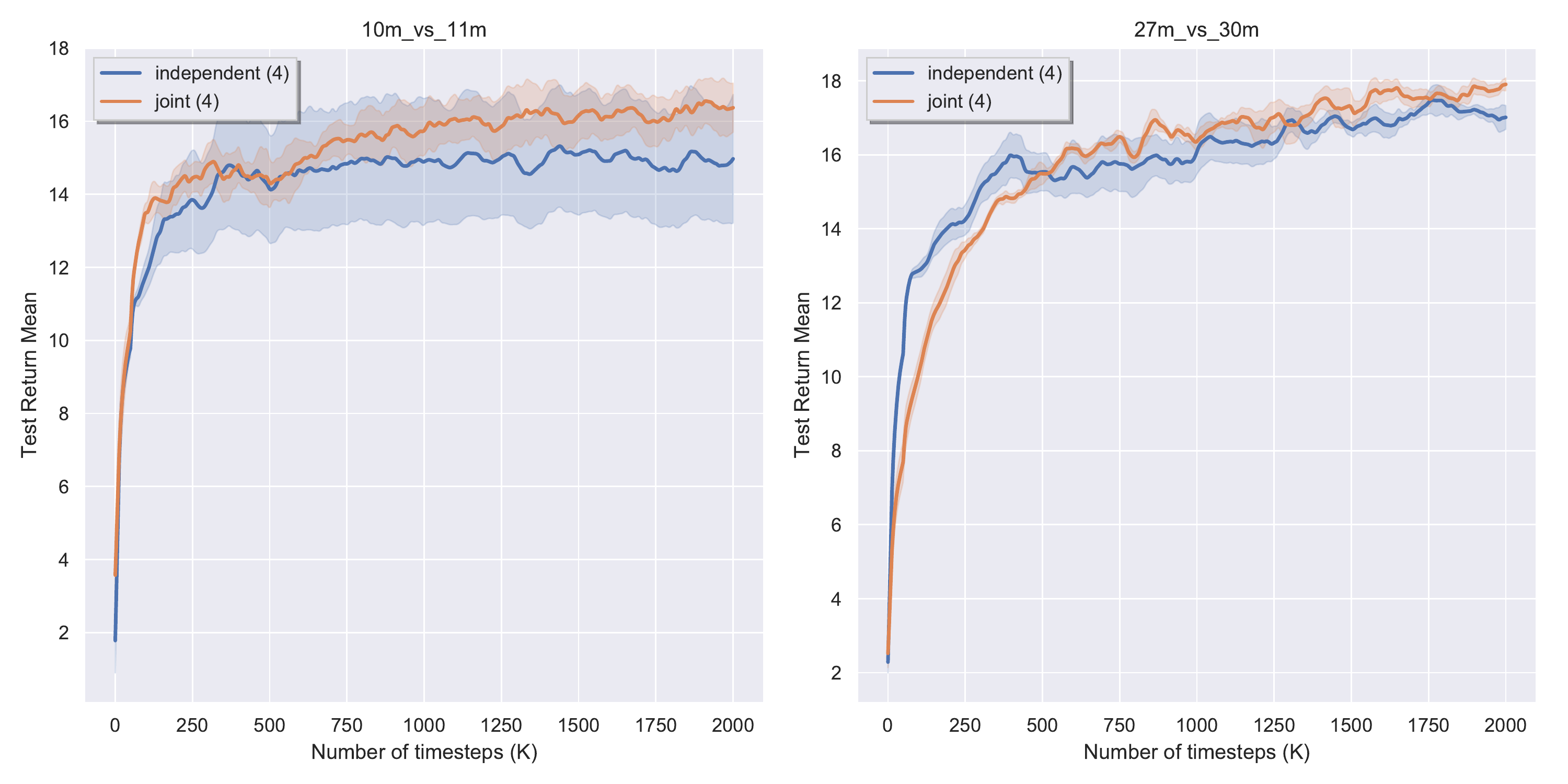}
   \includegraphics[width=0.40\linewidth]{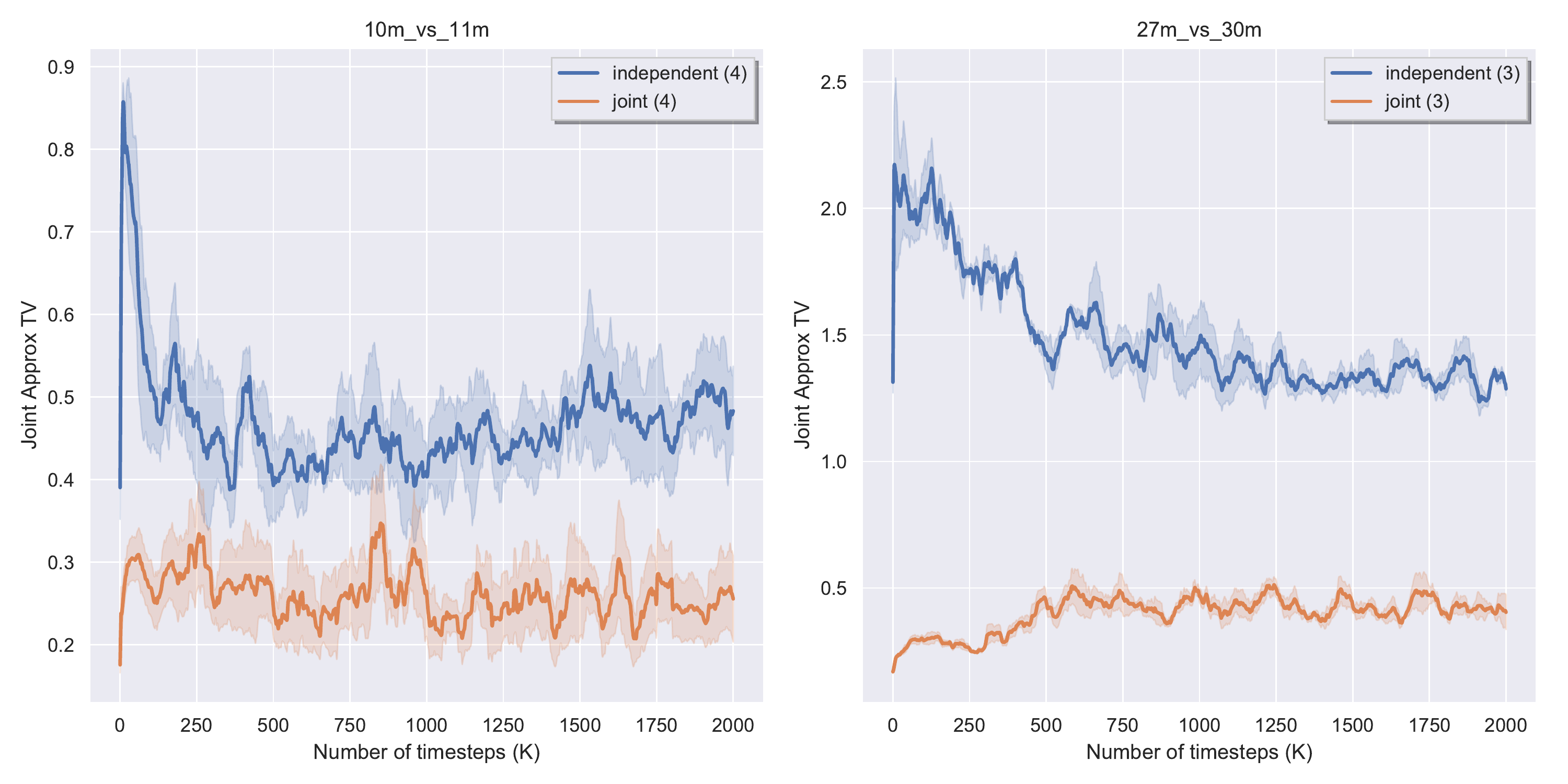}
   \includegraphics[width=0.40\linewidth]{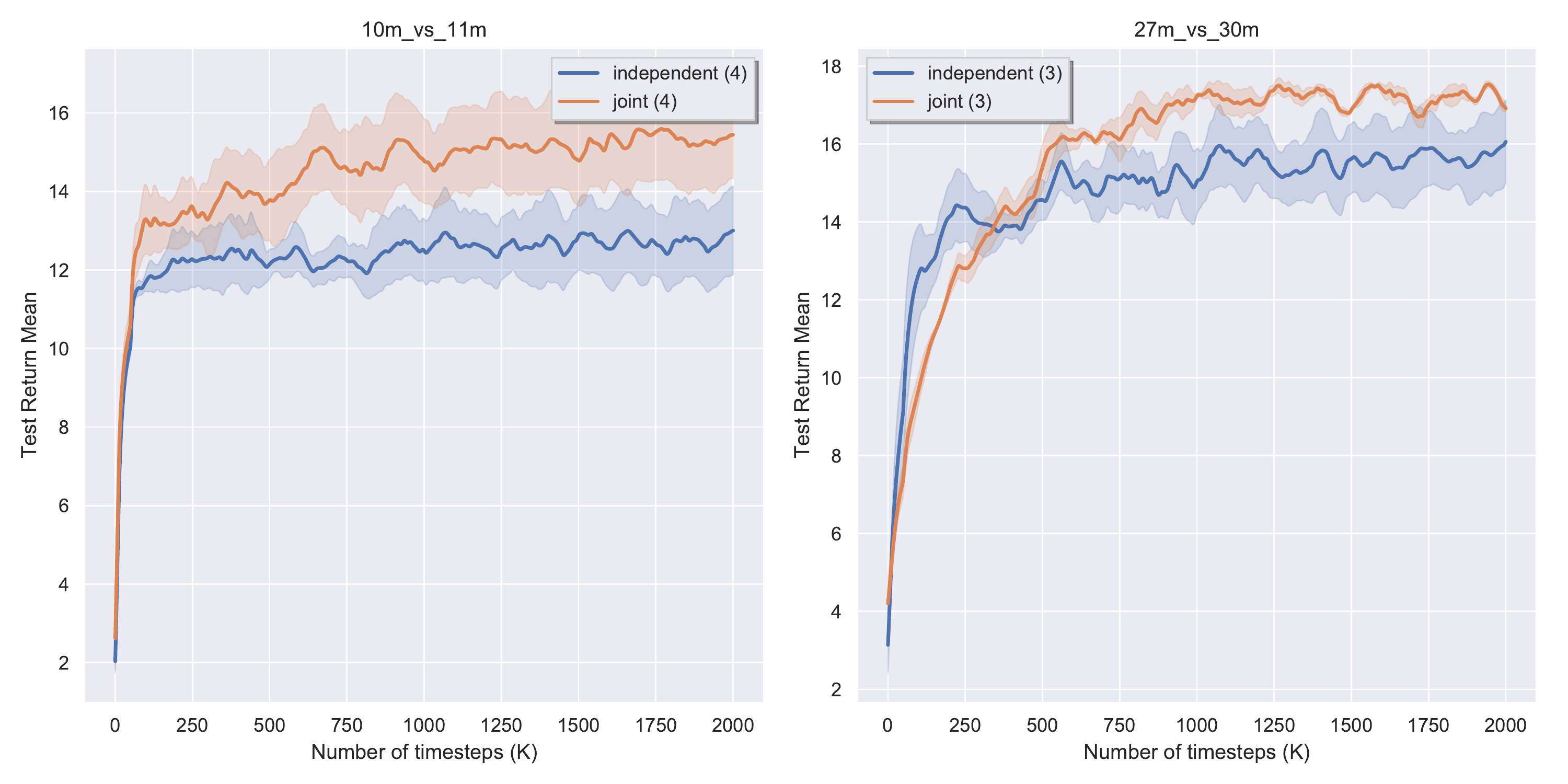}
   \caption{Joint divergence estimates and empirical returns for two types of ratio clipping at different clipping values: $0.1$ (first row), $0.3$ (second row) and $0.5$ (third row).}
   \label{fig:joint-independent-comparison-1}
\end{figure*}

\subsection{IPPO and MAPPO}
We show that the empirical performance of IPPO and MAPPO are very similar 
despite the fact that the advantage functions are learned differently. 
We evaluate IPPO and MAPPO on maps of varied difficulty and
numbers of agents. 
We heuristically set the clipping range based on the number of agents. 
Specifically, we set the clipping range $\epsilon$ for 
3s5z, 1c3s5z, 10m\_vs\_11m, and bane\_vs\_bane,  
as $0.1$, $0.1$, $0.1$, and $0.05$, respectively.
The results are presented in Figure~\ref{fig:ippo-mappo-smac-performance}. 
On the four maps considered, IPPO and MAPPO perform similarly. 
This phenomenon can be observed in~\cite{yu2021surprising}, 
which provides more comparisons between IPPO and MAPPO.
Such comparable performance also implies that,
for actor-critic methods in MARL, 
the way of training critics could be less crucial than enforcing the trust region constraint.

\subsection{Joint and Independent Ratio Clipping}
\begin{figure}
   \centering
   \includegraphics[width=0.90\linewidth]{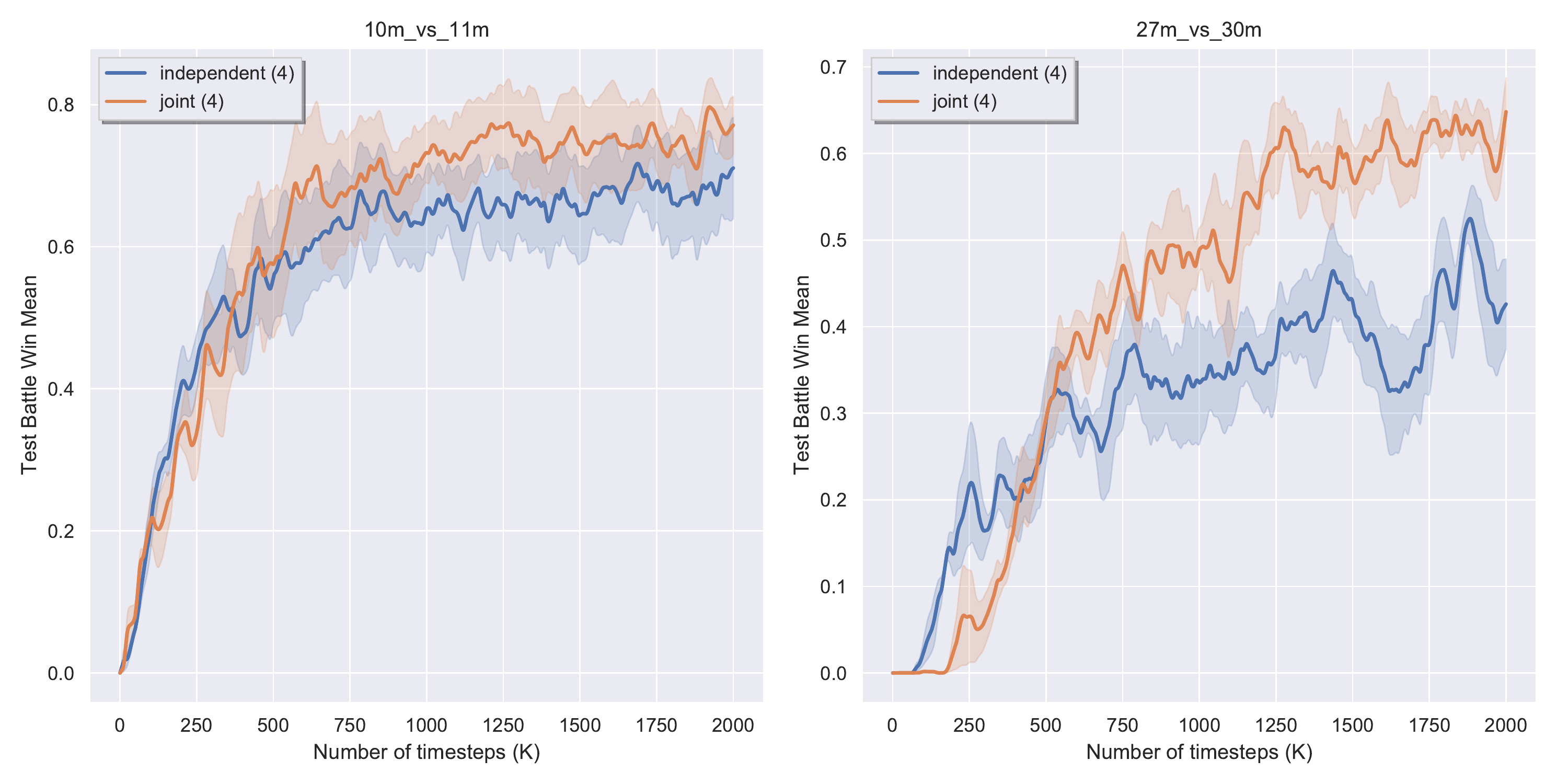}
   \includegraphics[width=0.90\linewidth]{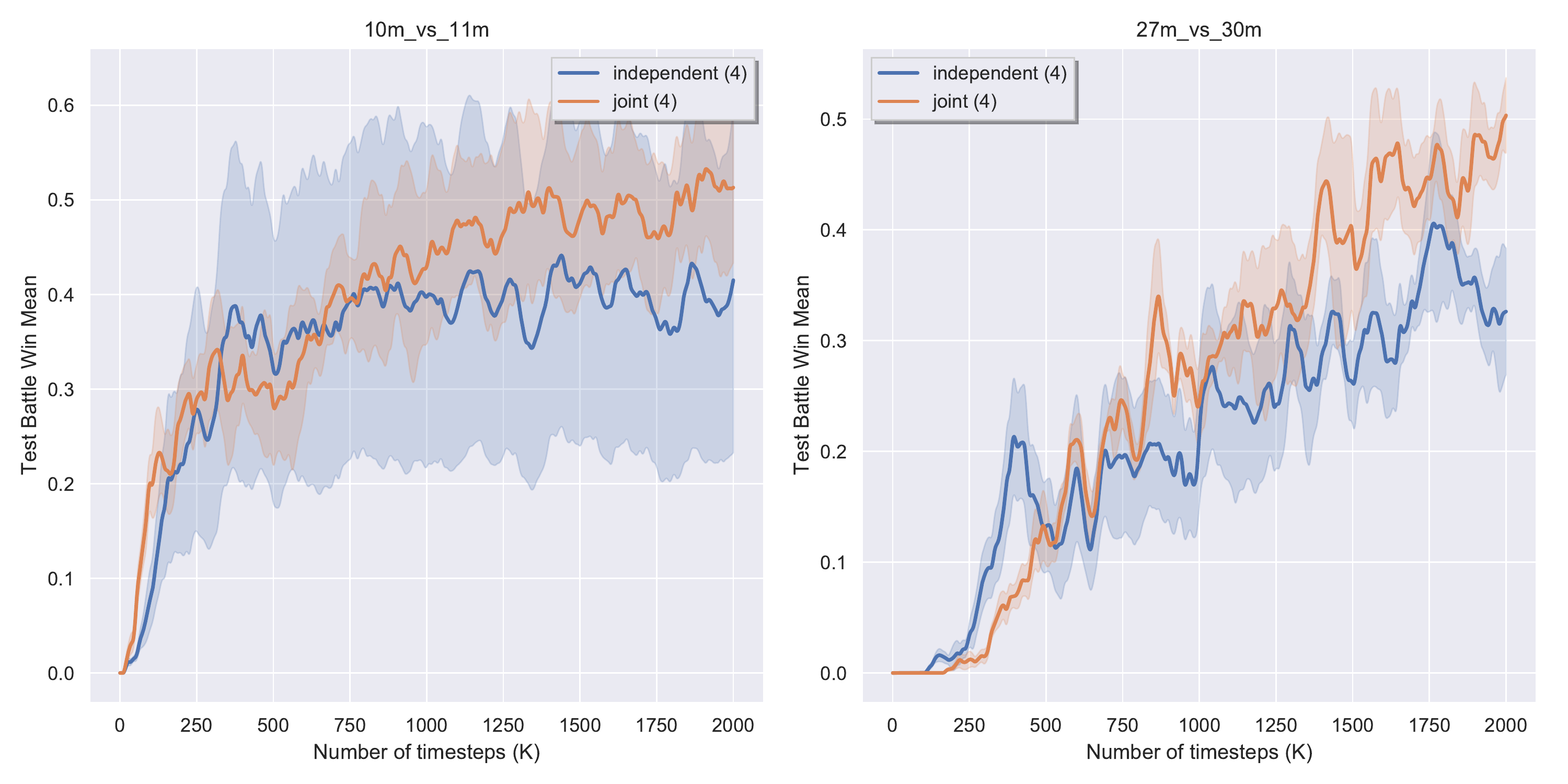}
   \includegraphics[width=0.90\linewidth]{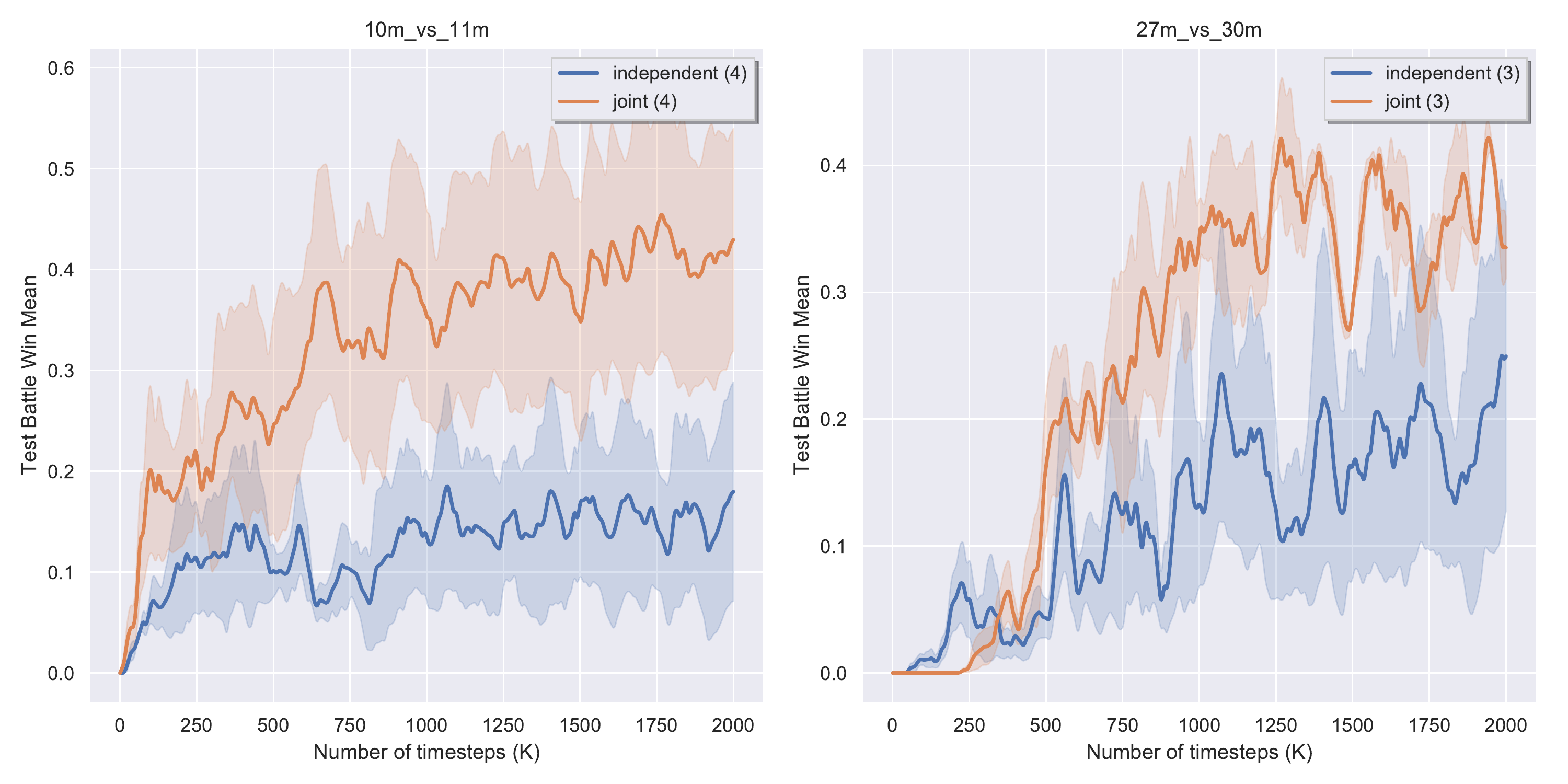}
   \caption{Test battle win rate for two types of ratio clipping at different clipping values: $0.1$ (first row), $0.3$ (second row) and $0.5$ (third row).}
   \label{fig:joint-independent-comparison-2}
\end{figure}

Finally, we apply the same clipping values to two types of clipping
(joint clipping and independent clipping), 
and use maps with many agents, i.e., 10m\_vs\_11m and 27m\_vs\_30m, 
to make the difference more salient (based on the theoretical results in the paper). 
The results are presented in Figure~\ref{fig:joint-independent-comparison-1} 
and~\ref{fig:joint-independent-comparison-2}. 

Compared to joint ratio clipping, the independent ratio clipping is 
more sensitive to the number of agents. 
In particular, for a small clipping value, e.g., $\epsilon=0.1$, 
joint ratio clipping consistently produces better performance than independent ratio clipping, 
even when the number of agents changes from 10 to 27. 
As the clipping value increases to $0.5$, the performance gap between these two types of clipping becomes larger, 
which is also aligned with our theoretical analysis.

\section{Related Work}
The use of trust region optimization in MARL traces back to 
parameter-sharing TRPO (PS-TRPO)~\cite{gupta2017cooperative}, 
which combines parameter sharing with TRPO  for cooperative multi-agent continuous control but provides no theoretical support.
Our analysis showing
that a trust region constraint is pivotal to guarantee performance improvement in MARL applies to PS-TRPO, among other algorithms.

Multi-agent trust region learning (MATRL)~\cite{wen2021game} uses a trust region for independent learning 
with a game-theoretical analysis in the policy space.
MATRL considers independent learning and proposes to enforce a trust region constraint 
by approximating the stable fixed point via a meta-game. 
Despite the improvement guarantee for joint policies, 
solving a meta-game itself can be challenging 
because its complexity  increases exponentially 
in the number of agents.
We instead consider  centralized learning and  
enforce the trust region constraint in a centralized and scalable way. 

Multi-Agent TRPO (MATRPO) directly extends TRPO to the multi-agent case~\cite{li2020multi} and divides the trust region by the number of agents.  However,
the analysis assumes a private reward for each agent, 
which yields different theoretical results from ours. 

Non-stationarity has been discussed in multi-agent mirror descent 
with trust region decomposition~\cite{li2021dealing}, 
which first decomposes the trust region for each decentralized policy 
and then approximates the KL divergence through additional training. 
However, this method needs to learn a fully centralized action-value function,
thus different from decentralized PPO algorithms.

One closely related work, Heterogeneous-Agent Trust Region Policy Optimization (HATPRO)~\cite{kuba2021trust}, 
shows that the joint advantage function in a Markov game can 
be decomposed as a summation of each agent's local advantages, 
from which a novel sequential policy update with monotonic improvement guarantee can be derived. 
We take a different perspective to show that non-stationarity of transition dynamics is decomposable and 
the derived monotonic improvement guarantee directly applies to IPPO and MAPPO. 

\balance

\section{Discussions}
\paragraph{Ratio Clipping}
Our analysis shows that using a small clipping value results in a small trust region, 
and thus small clipping values, e.g., $0.08$, $0.05$ and $0.03$, 
would be preferred for maps with a large number of agents, 
e.g., maps 10m\_vs\_11m (10 agents) and 27m\_vs\_30m (27 agents), 
see Figure~\ref{fig:small-clipping-values}.
On the other hand, if the clip value is too small, e.g., $\epsilon=0.01$ in maps with 5 and 8 agents,
the resultant trust region is also small and 
the update step in each iteration can thus be too small to effectively improve the policy. 
Furthermore, \cite{sun2022you} shows that the clipping 
may not necessarily bound the ratio ranges since such clipping depends on the learning rate and other compounding factors.
Alternatively, we can leverage the early stopping scheme, as suggested by~\cite{sun2022you}, 
to terminate the optimization epochs whenever the ratio deviations exceed a threshold. 
The early stopping has been reported to be more effective than ratio clipping. 
We leave the early stopping scheme in MARL as a future study since it is out of the scope of this paper. 

\paragraph{Independent Ratios vs. Joint Ratios}
While both independent and joint ratios enjoy theoretical guarantees, 
in theory, bounding joint ratios requires the advantage to be defined over joint actions, 
which could limits its application to a small number of agents. 
In contrast, bounding independent ratios has no such issue and may scale to large numbers of agents (if the trust region can be effectively enforced).
However, the SMAC results show that bounding joint ratios performs better than bounding independent ones even when the number of agents is large.
This could be due to that spreading trust region out evenly to each agent may not be an effective way to enforce the centralized trust region. 
It remains to an open question to find an optimal decomposition of trust region~\cite{li2020multi}. 

\paragraph{Centralized Value Functions} 
We show that both decentralized and decentralized advantage functions converge to the same fixed point, 
which, however, does not imply that the extra information have no impact on learning. 
In fact, the use of extra information can make the value learning easier for actor-critic methods~\cite{foerster_counterfactual_2017}. 
Also, as showed in~\cite{lyu2021contrasting}, 
the use of centralized critics or decentralized ones is a bias-variance trade off:
the centralized critic provides unbiased and correct on-policy return estimates,
while also introduce higher policy gradient variance than the decentralized critic in practice. 

\paragraph{Partial Observability} 
The theoretical analysis of this paper considers only DecMDPs. 
When the MDP is partial observable, 
we can leverage the neural architecture with memories to learn to memorize the observation history. 
We empirically evaluate the theoretical results in a partial observable domain, i.e., SMAC~\cite{samvelyan2019starcraft}, 
and used recurrent networks, i.e., LSTM, as the decentralized policy architecture 
to overcome any partial observability. 
These empirical results included in the paper corroborate our theoretical analysis.

\paragraph{Monotonic Improvement} 
It is worth noting that 
directly maximizing the lower bound in Theorem~\ref{theo:monotonic-decentralized} yields a monotonic guarantee. 
We alternatively consider using a hard constraint as an effective way to take large step-sizes. 
This change of optimization does not mean that Theorem~\ref{theo:monotonic-decentralized} is invalidated. 
In contrast, the TV in the constrained optimization is bounded to be small such that 
the performance of the updated policy can always be guaranteed, 
i.e., no policy collapse in training. 
Although our analysis may not seem critical from an algorithmic advancement perspective 
(since the existing IPPO and MAPPO implementations often achieve high scores on RL benchmarks), 
we believe that, besides algorithmic advancement for better performance, 
it is equally important to understand algorithms in a theoretically solid standpoint. 
Our results indeed elucidate IPPO and MAPPO in a principled perspective, and shed light on why existing algorithms work well.

\section{Conclusion}
In this paper, we presented a new monotonic improvement guarantee 
for optimizing decentralized policies in cooperative MARL. 
We showed that, despite the non-stationarity in IPPO and MAPPO,
a monotonic improvement guarantee still arises 
from enforcing the trust region constraint over all decentralized policies. 
This guarantee provides a theoretical understanding of the strong performance of IPPO and MAPPO.
Furthermore, we provided a theoretical foundation for proximal ratio clipping by
showing that a trust region constraint can be effectively enforced 
in a principled way by bounding independent ratios 
based on the number of agents in training. 
Finally, our empirical results supported the hypothesis 
that the strong performance of IPPO and MAPPO is a direct result of enforcing 
such a trust region via clipping in centralized training.

\begin{acks}
Mingfei Sun was partially supported by funding from Microsoft Research when this work was done. 
The experiments were made possible by a generous equipment grant from NVIDIA.
\end{acks}

\bibliographystyle{ACM-Reference-Format} 
\bibliography{ref}

\newpage
\onecolumn
\section{Appendix}

\subsection{Joint Ratio PPO}
\begin{algorithm}
\caption{Joint Ratio PPO (JR-PPO)}
\label{algo:joint-ratio}
\begin{algorithmic}
\FOR{iteration $i=0, 1, 2, \ldots$}
\STATE Roll out decentralized policies $[\pi_1, \pi_2, ..., \pi_N]$ in environment;
\STATE Compute centralized advantage estimates $A_{\vpi}(\vs, \va)$;
\STATE Compute joint ratios $\lambda_{\tilde{\pi}} = \frac{\tilde{\vpi}(\va|\vs)}{\vpi(\va|\vs)} = \prod_{k=1}^{N}\big[ \frac{\tilde{\pi}_k(a_k, s_k)}{\pi_k(a_k, s_k)} \big]$;
\STATE Optimize the surrogate objective $\max_{\tilde{\vpi}}\mathbb{E} \big[ \min\big(\lambda_{\tilde{\pi}} A_{\vpi}(\vs, \va), \clip(\lambda_{\tilde{\pi}}, 1\pm\epsilon) A_{\vpi}(\vs, \va)  \big) \big]$. 
\ENDFOR
\end{algorithmic}
\end{algorithm}

\subsection{Stationarity assumption in TRPO}\label{proof:stationarity-trpo}
The single-agent TRPO relies on the following analysis: 
\begin{align}
L_{\pi}(\tilde{\pi}) - L_{\pi}(\pi) &= \sum_{s} d(s)\sum_{a}\big(\tilde{\pi}(a|s) - \pi(a|s) \big) A_{\pi}(s, a) \\
&= \sum_{s} d(s)\sum_{a}\big(\tilde{\pi}(a|s) - \pi(a|s) \big) \big[ r(s) + \sum_{s^\prime}p(s^\prime|s, a)\gamma v_{\pi}(s^\prime) - v_{\pi}(s) \big] \\
&= \sum_{s} d(s)\sum_{s^\prime}\sum_{a}\big(\tilde{\pi}(a|s) - \pi(a|s) \big) p(s^\prime|s, a)\gamma v_{\pi}(s^\prime) \\
&= \sum_{s} d(s)\sum_{s^\prime}\sum_{a}\big(\tilde{\pi}(a|s)p(s^\prime|s, a) - \pi(a|s) p(s^\prime|s, a)\big)\gamma v_{\pi}(s^\prime).
\end{align}
This analysis is based on the assumption that $p(s^\prime|s, a)$ remains the same before and after $\pi$ is updated, 
such that transition shift $p_{\tilde{\pi}}(s^\prime|s) - p_{\pi}(s^\prime|s)$ is only caused by the agent's policy update, 
i.e., endogenously. 
Such analysis no longer holds when the transition dynamics $p(s^\prime|s, a)$ are non-stationary: 
$p_{\tilde{\pi}}(s^\prime|s, a)\neq p_{\pi}(s^\prime|s, a)$.

\subsection{Proofs}

\subsubsection{Proof of Proposition~\ref{prop:transition-shift}}
\begin{proof}\label{proof:transition-shift}
Assume agent $k$'s policy $\pi_k$ is executed independently of other agents policies $\pi_{-k}$, 
we have
\begin{align}
& \Delta^{\tilde{\pi}_1, ..., \tilde{\pi}_N}_{\pi_1, ..., \pi_N}(s'_k|s_k) \\
=& \sum_{\substack{s'_{-k}, s_{-k} \\ a_k, a_{-k}}} p(s'_k, s'_{-k}|s_k, s_{-k}, a_k, a_{-k})\big[ \tilde{\pi}_k(a_k|s_k)\tilde{\pi}_{-k}(a_{-k}|s_{-k}) - \pi_{k}(a_k|s_k)\pi_{-k}(a_{-k}|s_{-k})\big]\\
=& \sum_{\substack{s'_{-k}, s_{-k} \\ a_k, a_{-k}}} p(s'_k, s'_{-k}|s_k, s_{-k}, a_k, a_{-k})\cdot
\big[ \underbrace{\tilde{\pi}_{k}(a_{k}|s_{k})\tilde{\pi}_{-k}(a_{-k}|s_{-k}) - \tilde{\pi}_k(a_k|s_k)\pi_{-k}(a_{-k}|s_{-k})}_{\text{exogenous}} \\
& + \underbrace{\tilde{\pi}_k(a_k|s_k)\pi_{-k}(a_{-k}|s_{-k}) - \pi_{k}(a_k|s_k)\pi_{-k}(a_{-k}|s_{-k})}_{\text{endogenous}}\big].
\end{align}
The above decomposition can be repeated such that the exogenous part 
can be translated into endogenous parts that are specific to each agent.
Specifically, repeat the decomposition for the exogenous part by considering 
agent $k'$ ($k'\neq k$):
\begin{align}
&\tilde{\pi}_k(a_k|s_k)\tilde{\pi}_{-k}(a_{-k}|s_{-k}) - \tilde{\pi}_k(a_k|s_k)\pi_{-k}(a_{-k}|s_{-k}) \\
=& \tilde{\pi}_k(a_k|s_k)\left[\tilde{\pi}_{k'}(a_{k'}|s_{k'})\tilde{\pi}_{-\{k,k'\}}(a_{-\{k,k'\}}|s_{-\{k,k'\}}) - \pi_{k'}(a_{k'}|s_{k'})\pi_{-\{k,k'\}}(a_{-\{k,k'\}}|s_{-\{k,k'\}})\right] \\
=& \tilde{\pi}_k(a_k|s_k)\Big[\underbrace{\tilde{\pi}_{k'}(a_{k'}|s_{k'})\tilde{\pi}_{-\{k,k'\}}(a_{-\{k,k'\}}|s_{-\{k,k'\}}) - \tilde{\pi}_{k'}(a_{k'}|s_{k'})\pi_{-\{k,k'\}}(a_{-\{k,k'\}}|s_{-\{k,k'\}})}_{\pi_k\text{-exogenous}} \\
& + \underbrace{\tilde{\pi}_{k'}(a_{k'}|s_{k'})\pi_{-\{k,k'\}}(a_{-\{k,k'\}}|s_{-\{k,k'\}}) - \pi_{k'}(a_{k'}|s_{k'})\pi_{-\{k,k'\}}(a_{-\{k,k'\}}|s_{-\{k,k'\}})}_{\pi_k\text{-endogenous}} \Big]. 
\end{align}
So on and so forth, one can decompose 
$\Delta^{\tilde{\pi}_1, ..., \tilde{\pi}_N}_{\pi_1, ..., \pi_N}(s'_k|s_k)$ as follows:
\begin{equation}
\Delta^{\tilde{\pi}_1, ..., \tilde{\pi}_N}_{\pi_1, ..., \pi_N}(s'_k|s_k) = \Delta^{\tilde{\pi}_1,\pi_2, ..., \pi_N}_{\pi_1,\pi_2, ..., \pi_N}(s'_k|s_k) + \Delta^{\tilde{\pi}_1, \tilde{\pi}_2,\pi_3,..., \pi_N}_{\tilde{\pi}_1,\pi_2,\pi_3, ..., \pi_N}(s'_k|s_k) +...+ \Delta^{\tilde{\pi}_1, ..., \tilde{\pi}_{N-1},\tilde{\pi}_N}_{\tilde{\pi}_1, ..., \tilde{\pi}_{N-1},\pi_N}(s'_k|s_k),
\end{equation}
which implies that the state transition shift at local observation $s_k$ is caused 
by the shifts arising from all decentralized policies. 
\end{proof}

\subsubsection{Proof of Theorem~\ref{theo:monotonic-decentralized}}
\begin{proof}\label{proof:monotonic-decentralized}
Define the discounted state distribution $d_{\tilde{\pi}_k}$ for $\tilde{\pi}_k$ as 
\begin{equation}
    d_{\tilde{\pi}_k}(s_k) \triangleq (1-\gamma) \sum_{t=0}^{\infty} \gamma^t \cdot \text{Probability}\big(S^{[t]}_k=s_k|\tilde{\pi}_k \big). 
\end{equation}
If $p_0(s_k)$ is the distribution of starting states ($p_0(s_k)>0, \forall s_k\in\mathcal{S}_k$) and $p_{\tilde{\pi}_k}(s_k^\prime|s_k, a_k)$ is the stationary transition model, then 
\begin{equation}
d_{\tilde{\pi}_k}(s^\prime_k) = (1-\gamma) p_0(s^\prime_k) + \gamma\sum_{s_k, a_k}p_{\tilde{\pi}_k}(s^\prime_k|s_k, a_k)\tilde{\pi}(a_k|s_k)d_{\tilde{\pi}_k}(s_k)
\end{equation}
In vector notation, 
\begin{equation}
D_{\tilde{\pi}_k} = (1-\gamma) P_0 + \gamma D_{\tilde{\pi}_k} P_{\tilde{\pi}_k} 
\end{equation}
Thus, 
\begin{equation}
(1-\gamma) P_0 \left(I -\gamma P_{\tilde{\pi}_k}\right)^{-1} = D_{\tilde{\pi}_k}. 
\end{equation}
We also have
\begin{align}
V_{\tilde{\pi}_k} - V_{\pi_k} &= R_{\tilde{\pi}_k} + \gamma P_{\tilde{\pi}_k} V_{\tilde{\pi}_k} - R_{\pi_k} - \gamma P_{\pi_k} V_{\pi_k} \\
&= \gamma P_{\tilde{\pi}_k} V_{\tilde{\pi}_k} - \gamma P_{\tilde{\pi}_k} V_{\pi_k} + \gamma P_{\tilde{\pi}_k} V_{\pi_k} - \gamma P_{\pi_k} V_{\pi_k} \\
&= \gamma P_{\tilde{\pi}_k} \left( V_{\tilde{\pi}_k} - V_{\pi_k} \right) + \gamma P_{\tilde{\pi}_k} V_{\pi_k} - \gamma P_{\pi_k} V_{\pi_k}, 
\end{align}
i.e., 
\begin{align}
\left( I - \gamma P_{\tilde{\pi}_k}\right) \left(V_{\tilde{\pi}_k} - V_{\pi_k}\right) = \gamma P_{\tilde{\pi}_k} V_{\pi_k} - \gamma P_{\pi_k} V_{\pi_k}. 
\end{align}
Thus, 
\begin{align}
V_{\tilde{\pi}_k} - V_{\pi_k} = \left( I - \gamma P_{\tilde{\pi}_k}\right)^{-1} \left(\gamma P_{\tilde{\pi}_k} V_{\pi_k} - \gamma P_{\pi_k} V_{\pi_k}\right)
\end{align}
Therefore, based on the definition of $J(\pi_k)$: $J(\pi_k) \triangleq \mathbb{E}_{s_0\sim p_0}[v_{\pi_k}(s_0)]$, 
we have
\begin{align}
J(\tilde{\pi}_k) - J(\pi_k) &= \mathbb{E}_{s_k^{[0]}\sim p_0}[v_{\tilde{\pi}_k}(s_k) - v_{\pi_k}(s_k)] \\
&= P_0\left( I - \gamma P_{\tilde{\pi}_k}\right)^{-1} \left(\gamma P_{\tilde{\pi}_k} V_{\pi_k} - \gamma P_{\pi_k} V_{\pi_k}\right) \\
&= \frac{1}{1-\gamma} D_{\tilde{\pi}_k}\left(\gamma P_{\tilde{\pi}_k} V_{\pi_k} - \gamma P_{\pi_k} V_{\pi_k}\right) \\
&= \frac{1}{1-\gamma} \sum_{s_k}d_{\tilde{\pi}_k}(s_k)\left[ \gamma \sum_{s'_k}\Delta_{\pi_1, \pi_2, ..., \pi_N}^{\tilde{\pi}_1, \tilde{\pi}_2, ..., \tilde{\pi}_N}(s'_k|s_k) v_{\pi_k}(s'_k)  \right]
\end{align}
Proposition~\ref{prop:transition-shift} suggests that 
\begin{equation}
\Delta_{\pi_1, \pi_2, ..., \pi_N}^{\tilde{\pi}_1, \tilde{\pi}_2, ..., \tilde{\pi}_N}(s_k^\prime | s_k) = 
\Delta_{\pi_1, \pi_2, ..., \pi_N}^{\tilde{\pi}_1, \pi_2, ...,\pi_N}(s_k^\prime | s_k) 
+\Delta_{\tilde{\pi}_1, \pi_2, \pi_3, ..., \pi_N}^{\tilde{\pi}_1, \tilde{\pi}_2,\pi_3, ...,\pi_N}(s_k^\prime | s_k)
+ \dots +\Delta_{\tilde{\pi}_1, \tilde{\pi}_2, ..., \tilde{\pi}_{N-1}, \pi_N}^{\tilde{\pi}_1, \tilde{\pi}_2, ..., \tilde{\pi}_{N-1}, \tilde{\pi}_N}(s_k^\prime | s_k). 
\end{equation}
Thus, 
\begin{multline}
  \sum_{s'_k}\Delta_{\pi_1, \pi_2, ..., \pi_N}^{\tilde{\pi}_1, \tilde{\pi}_2, ..., \tilde{\pi}_N}(s'_k|s_k) \gamma v_{\pi_k}(s'_k)
  = \sum_{s_k^\prime} \bigg( 
  \Delta_{\pi_1, \pi_2, ..., \pi_N}^{\tilde{\pi}_1, \pi_2, ...,\pi_N}(s_k^\prime | s_k) + \\
  \Delta_{\tilde{\pi}_1, \pi_2, \pi_3, ..., \pi_N}^{\tilde{\pi}_1, \tilde{\pi}_2,\pi_3, ...,\pi_N}(s_k^\prime | s_k)
  +\Delta_{\tilde{\pi}_1, \tilde{\pi}_2, \pi_3, \pi_4, ..., \pi_N}^{\tilde{\pi}_1, \tilde{\pi}_2, \tilde{\pi}_3, \pi_4, ..., \pi_N}(s_k^\prime | s_k) 
  + \dots +\Delta_{\tilde{\pi}_1, \tilde{\pi}_2, ..., \tilde{\pi}_{N-1}, \pi_N}^{\tilde{\pi}_1, \tilde{\pi}_2, ..., \tilde{\pi}_{N-1}, \tilde{\pi}_N}(s_k^\prime | s_k) 
  \bigg) \gamma v_{\pi_k}(s_k^\prime)
\end{multline}

For one of these summation terms, we have the following
\begin{align}
   &\sum_{s^\prime_k} \Big[ \Delta^{\tilde{\pi}_1,...,\tilde{\pi}_{j-1},\tilde{\pi}_j, ..., \pi_N}_{\tilde{\pi}_1,...,\tilde{\pi}_{j-1},\pi_j,..., \pi_N}(s_k^\prime|s_k) \Big]\gamma v_{\pi_k}(s^\prime_k) \\
   = & \sum_{s^\prime_k} \sum_{a_k} \Big( p_{\tilde{\pi}_1,...,\tilde{\pi}_{j-1},\pi_j,..., \pi_N}(s^\prime_k|s_k, a_k)\tilde{\pi}_j(a_k|s_k) - p_{\tilde{\pi}_1,...,\tilde{\pi}_{j-1},\pi_j,..., \pi_N}(s^\prime_k|s_k, a_k)\pi_j(a_k|s_k) \Big)\gamma v_{\pi_k}(s^\prime_k) \\
   = & \sum_{a_k} \big( \tilde{\pi}_j(a_k|s_k) - \pi_j(a_k|s_k) \big)\sum_{s^\prime_k} p_{\tilde{\pi}_1,...,\tilde{\pi}_{j-1},\pi_j,..., \pi_N}(s^\prime_k|s_k, a_k)\gamma v_{\pi_k}(s^\prime_k) \\
   = & \sum_{a_k} \big( \tilde{\pi}_j(a_k|s_k) - \pi_j(a_k|s_k) \big)\big[r(s_k) + \sum_{s^\prime_k} p_{\tilde{\pi}_1,...,\tilde{\pi}_{j-1},\pi_j,..., \pi_N}(s^\prime_k|s_k, a_k)\gamma v_{\pi_k}(s^\prime_k) - v_{\pi_{k}}(s_k)\big] \\
   \intertext{\red{\centering $\sum_{a_k} (\tilde{\pi}(...)-\pi(...)[r(s_k) + v_{\pi_k}(s_k)])=0$ because $\sum_{a_k} (\tilde{\pi}(...)-\pi(...))=0$ and $r(s_k)$, $v_{\pi_k}(s_k)$ are both independent of $a_k$;}}
   = & \sum_{a_k} \big( \tilde{\pi}_j(a_k|s_k) - \pi_j(a_k|s_k) \big) A_{\pi_k}^{\pi_j}(s_k, a_k) 
\end{align}
The last transition is based on the following definition:
\begin{equation}
A_{\pi_k}^{\pi_j}(s_k, a_k) \triangleq r(s_k) + \gamma\sum_{s^\prime_k} p_{\tilde{\pi}_1,...,\tilde{\pi}_{j-1},\pi_j,..., \pi_N}(s^\prime_k|s_k, a_k) v_{\pi_k}(s^\prime_k) - v_{\pi_{k}}(s_k), 
\end{equation}
Thus, 
\begin{align}
  & J(\tilde{\pi}_k) - J(\pi_k) \\
= & \frac{1}{1-\gamma} \sum_{s_k}d_{\tilde{\pi}_k}(s_k)\left[ \sum_{j=1}^{N} \sum_{a_k} \big( \tilde{\pi}_j(a_k|s_k) - \pi_j(a_k|s_k) \big) A_{\pi_k}^{\pi_j}(s_k, a_k) \right] \\
= & \frac{1}{1-\gamma} \sum_{j=1}^{N} \mathbb{E}_{ \blue{s_k\sim d_{\tilde{\pi}_k}} } \mathbb{E}_{a_k\sim\pi_j} \Big[\frac{\tilde{\pi}_j(a_k|s_k)}{\pi_j(a_k|s_k)}  - 1  \Big] A_{\pi_k}^{\pi_j}(s_k, a_k) \\
= & \frac{1}{1-\gamma} \sum_{j=1}^{N} \left[ \underbrace{\mathbb{E}_{ \red{s_k\sim d_{\pi_k}} } \mathbb{E}_{a_k\sim\pi_j} \Big[\frac{\tilde{\pi}_j(a_k|s_k)}{\pi_j(a_k|s_k)}  - 1  \Big] A_{\pi_k}^{\pi_j}(s_k, a_k)}_{\text{Surrogate term}} + \underbrace{\mathbb{E}_{ \blue{s_k\sim d_{\tilde{\pi}_k}} } \mathbb{E}_{a_k\sim\pi_j} \Big[\frac{\tilde{\pi}_j(a_k|s_k)}{\pi_j(a_k|s_k)}  - 1  \Big] A_{\pi_k}^{\pi_j}(s_k, a_k) - \mathbb{E}_{ \red{s_k\sim d_{\pi_k}} } \mathbb{E}_{a_k\sim\pi_j} \Big[\frac{\tilde{\pi}_j(a_k|s_k)}{\pi_j(a_k|s_k)}  - 1  \Big] A_{\pi_k}^{\pi_j}(s_k, a_k)}_{\text{Correction term}}  \right]
\end{align}

The surrogate term is the objective to maximize. 
Now, we consider the correction term. 
\begin{multline}
  \mathbb{E}_{ \blue{s_k\sim d_{\tilde{\pi}_k}} } \mathbb{E}_{a_k\sim\pi_j} \Big[\frac{\tilde{\pi}_j(a_k|s_k)}{\pi_j(a_k|s_k)}  - 1  \Big] A_{\pi_k}^{\pi_j}(s_k, a_k) - \mathbb{E}_{ \red{s_k\sim d_{\pi_k}} } \mathbb{E}_{a_k\sim\pi_j} \Big[\frac{\tilde{\pi}_j(a_k|s_k)}{\pi_j(a_k|s_k)}  - 1  \Big] A_{\pi_k}^{\pi_j}(s_k, a_k) = \\
= \sum_{s_k} \big[\red{d_{\tilde{\pi}_k}(s_k)} - \blue{d_{\pi_k}(s_k)}\big] \underbrace{\sum_{a_k}\big[\tilde{\pi}_j(a_k|s_k) - \pi_j(a_k|s_k) \big] A_{\pi_k}^{\pi_j}(s_k, a_k)}_{\text{Denoted as $A_{\tilde{\pi}_k}$ }}
\end{multline}
Using vector notation $\red{d_{\tilde{\pi}_k}} - \blue{d_{\pi_k}}$ and $A_{\tilde{\pi}_k}$, 
The above term is bounded by applying Holder's inequality: for any $p, q\in[1, \infty]$, 
such that $\frac{1}{p} + \frac{1}{q} = 1$, 
we have
\begin{equation}
\norm{(\red{d_{\tilde{\pi}_k}} - \blue{d_{\pi_k}})\cdot A_{\tilde{\pi}_k} }_1 \leq \norm{\red{d_{\tilde{\pi}_k}} - \blue{d_{\pi_k}}}_p \norm{A_{\tilde{\pi}_k}}_q. 
\end{equation}
Consider the case $p=1$ and $q=\infty$ as in~\cite{schulman2015trust} and~\cite{achiam2017constrained}, 
and aim at bounding $ \norm{\red{d_{\tilde{\pi}_k}} - \blue{d_{\pi_k}}}_1$ and $\norm{A_{\tilde{\pi}_k}}_\infty$. 

We first show how to bound $\norm{\red{d_{\tilde{\pi}_k}} - \blue{d_{\pi_k}}}_1$.

Let $G^{s_k} = (1+\gamma P_{\pi_k}^{s_k} + (\gamma P_{\pi_k}^{s_k})^2 + ... = (1-\gamma P_{\pi_k}^{s_k})^{-1}$  
and $\tilde{G}^{s_k} = (1+\gamma P_{\tilde{\pi}_k}^{s_k} + (\gamma P_{\tilde{\pi}_k}^{s_k})^2 + ... = (1-\gamma P_{\tilde{\pi}_k}^{s_k})^{-1}$ denote the distribution of state $s_k$ under $\pi_k$ and $\tilde{\pi}_k$. 
We will use the convention that $d$ (a density on state space) is a vector and $r$ (a reward function on state space) is a dual vector (i.e., linear functional on vectors), thus $r d$ is a scalar meaning the expected reward under density $d$. 
Note that $J(\pi)=rG d_0$, and $J(\tilde{\pi})=r\tilde{G} d_0$. 
Note $\Delta_{\pi_k}^{\tilde{\pi}_k} \triangleq P_{\tilde{\pi}_k}^{s_k} - P_{\pi_k}^{s_k}$. 
Using the perturbation theory, we have the following 
\begin{equation}
[G^{s_k}]^{-1} - [\tilde{G}^{s_k}]^{-1} =\gamma P_{\tilde{\pi}_k}^{s_k} - \gamma P_{\pi_k}^{s_k} = \gamma \Delta_{\pi_k}^{\tilde{\pi}_k}. 
\end{equation}
Right multiply by $G^{s_k}$ and left multiply by $\tilde{G}^{s_k}$: 
\begin{equation}
\tilde{G}^{s_k} - G^{s_k} = \gamma \tilde{G}^{s_k}\Delta_{\pi_k}^{\tilde{\pi}_k} G^{s_k}. 
\end{equation}
Thus, 
\begin{equation}
\red{d_{\tilde{\pi}_k}} - \blue{d_{\pi_k}} 
= (1-\gamma)(\tilde{G}^{s_k} - G^{s_k})p_0 = (1-\gamma) \gamma \tilde{G}^{s_k}\Delta_{\pi_k}^{\tilde{\pi}_k} G^{s_k} p_0 
= \gamma \tilde{G}^{s_k}\Delta_{\pi_k}^{\tilde{\pi}_k} \blue{d_{\pi_k}}. 
\end{equation}
According to the $l_1$ operator norm
$\norm{A}_1 = \sup_{ d} \left\{ \frac{\norm{A d}_1}{\norm{ d}_1} \right\}$, 
we have
\begin{align}
\norm{\red{d_{\tilde{\pi}_k}} - \blue{d_{\pi_k}}}_1 &= \gamma \norm{ \tilde{G}^{s_k}\Delta_{\pi_k}^{\tilde{\pi}_k} \blue{d_{\pi_k}} }_1 \\
&\leq \gamma \norm{ \tilde{G}^{s_k}}_1 \cdot \norm{\Delta_{\pi_k}^{\tilde{\pi}_k} \blue{d_{\pi_k}} }_1 \\
&= \gamma \norm{ (\mathbf{1} + \gamma P_{\tilde{\pi}_k} + (\gamma P_{\tilde{\pi}_k})^2 + ...)}_1 \cdot \norm{\Delta_{\pi_k}^{\tilde{\pi}_k} \blue{d_{\pi_k}} }_1 \\
&\leq \gamma (1 + \gamma \norm{P_{\tilde{\pi}_k}}_1 + \gamma^2 \norm{P_{\tilde{\pi}_k}^2}_1 + ...) \cdot \norm{\Delta_{\pi_k}^{\tilde{\pi}_k} \blue{d_{\pi_k}} }_1 \\
&= \frac{\gamma}{1-\gamma} \norm{\Delta_{\pi_k}^{\tilde{\pi}_k} \blue{d_{\pi_k}} }_1
\end{align}
Meanwhile, 
\begin{align}
  \norm{ \Delta_{\pi_k}^{\tilde{\pi}_k} \blue{d_{\pi_k}} }_1 &= \sum_{s^\prime_k} \left| \sum_{s_k} \blue{d_{\pi_k}(s_k)} \big( p_{\tilde{\pi}_1, \tilde{\pi}_2, ...,\tilde{\pi}_k, ..., \tilde{\pi}_N}(s_k^\prime|s_k) - p_{\pi_1, \pi_2, ..., \pi_k, ..., \pi_N}(s_k^\prime|s_k) \big) \right| \\
  &\leq \sum_{s_k} \blue{d_{\pi_k}(s_k)} \sum_{s^\prime_k} \left| p_{\tilde{\pi}_1, \tilde{\pi}_2, ...,\tilde{\pi}_k, ..., \tilde{\pi}_N}(s_k^\prime|s_k) - p_{\pi_1, \pi_2, ..., \pi_k, ..., \pi_N}(s_k^\prime|s_k) \right| \\
  &\leq \sum_{s_k} \blue{d_{\pi_k}(s_k)} \sum_{i=1}^{N} \sum_{a_k} \left| \tilde{\pi}_i(a_k | s_k) - \pi_i(a_k|s_k) \right| \\
  &= \sum_{s_k, a_k} \blue{d_{\pi_k}(s_k)} \sum_{i=1}^{N} \pi_i(a_k|s_k) \left| \frac{\tilde{\pi}_i(a_k | s_k)}{\pi_i(a_k|s_k)}  - 1  \right| \\
  &= \sum_{s_k, a_k} \blue{d_{\pi_k}(s_k)} \sum_{i=1}^{N} 2\cdot \TV\big(\pi_i(\cdot|s_k), \tilde{\pi}_i(\cdot|s_k) \big)
\end{align}
Thus, $\norm{\red{d_{\tilde{\pi}_k}} - \blue{d_{\pi_k}}}_1 \leq \frac{2\gamma}{1-\gamma} \mathbb{E}_{s\sim d_{\pi_k}}[\sum_{i=1}^{N} \TV\big(\pi_i(\cdot|s_k), \tilde{\pi}_i(\cdot|s_k) \big)]$.

Next, we show how to bound $\norm{A_{\tilde{\pi}_k}}_\infty$. 
\begin{equation}
\norm{A_{\tilde{\pi}_k}}_\infty = \max_{s_k, a_k} \left|\sum_{a_k} \big[ \tilde{\pi}_k(a_k|s_k) - \pi_k(a_k|s_k)\big] A_{\pi_k}^{\pi_j}(s_k, a_k) \right| \leq \max_{s_k, a_k} \left|A_{\pi_k}^{\pi_j}(s_k, a_k)\right| = \xi. 
\end{equation}

Combined, we have
\begin{equation}
J(\tilde{\pi}_k) - J(\pi_k) \geq \frac{1}{1-\gamma}\sum^N_{j=1} \bigg\{  L_{\pi_k}(\tilde{\pi}_j) - \frac{2\gamma\xi\alpha}{1-\gamma} \bigg\} = \frac{1}{1-\gamma} \bigg\{  \sum^N_{j=1}L_{\pi_k}(\tilde{\pi}_j) - \frac{2N\gamma\xi\alpha}{1-\gamma} \bigg\}, 
\end{equation}
which concludes the proof. 
\end{proof}

\subsubsection{Proof of Proposition~\ref{theo:trust-region-clipping}}
\begin{proof}\label{proof:trust-region-clipping}
For $\TV$ divergence, we have
$\TV(\mu(x), \nu(x))=\sum_{\mu(x)>\nu(x)}[\mu(x)-\nu(x)]$ 
where $\mu$ and $\nu$ are two distributions. 
Thus,
\begin{align}
  & \mathbb{E}_{s\sim d_{\pi_k}}\big[\TV(\pi_k, \tilde{\pi}_k)\big] \\
= & \mathbb{E}_{s\sim d_{\pi_k}}\sum_{\substack{a_k \text{ s.t. }\\ \tilde{\pi}_j(a_k|s_k)\geq \pi_j(a_k|s_k)}}[\tilde{\pi}_j(a_k|s_k) - \pi_j(a_k|s_k)] \\
= & \mathbb{E}_{s\sim d_{\pi_k}}\sum_{\substack{a_k \text{ s.t. }\\ \tilde{\pi}_j(a_k|s_k)\geq \pi_j(a_k|s_k)}} \pi_j(a_k|s_k) \left[\frac{\tilde{\pi}_j(a_k|s_k)}{\pi_j(a_k|s_k)} - 1 \right] \\
\leq & \mathbb{E}_{s\sim d_{\pi_k}}\sum_{\substack{a_k\text{ s.t. } \\ \tilde{\pi}_j(a_k|s_k)\geq \pi_j(a_k|s_k)}}[\epsilon_j\pi_j(a_k|s_k)] 
\leq \epsilon_j. 
\end{align}
Also, 
\begin{align}
  & \mathbb{E}_{s\sim d_{\pi_k}}\big[\TV(\pi_k, \tilde{\pi}_k)\big] \\
= & \mathbb{E}_{s\sim d_{\pi_k}}\sum_{\substack{a_k \text{ s.t. }\\ \tilde{\pi}_j(a_k|s_k)\leq \pi_j(a_k|s_k)}}[\pi_j(a_k|s_k) - \tilde{\pi}_j(a_k|s_k)] \\
= & \mathbb{E}_{s\sim d_{\pi_k}}\sum_{\substack{a_k \text{ s.t. }\\ \tilde{\pi}_j(a_k|s_k)\leq \pi_j(a_k|s_k)}} \tilde{\pi}_j(a_k|s_k) \left[\frac{\pi_j(a_k|s_k)}{\tilde{\pi}_j(a_k|s_k)} - 1 \right] \\
\leq & \mathbb{E}_{s\sim d_{\pi_k}}\sum_{\substack{a_k\text{ s.t. } \\ \tilde{\pi}_j(a_k|s_k)\leq \pi_j(a_k|s_k)}}[\epsilon_j\tilde{\pi}_j(a_k|s_k)] 
\leq \epsilon_j. 
\end{align}
As $\TV$ is a bounded divergence between $[0, 1]$, 
the ratio guarantee makes sense when $\epsilon_j \leq 1.0$. 
\end{proof}

\subsection{Experiment details and more results}\label{appen:more-exps}
The number of agents in each is given in Table~\ref{tab:number-of-agents}. 
\begin{table}[h]
  \centering
  \caption{Number of agents on maps.}
  \begin{tabular}{cc}\\\toprule  
  SMAC Map & Number of agents \\\midrule
  2s\_vs\_1sc     &2 \\  \midrule
  3s\_vs\_5z      &3 \\  \midrule
  2s3z            &5 \\  \midrule
  6h\_vs\_8z      &6 \\  \midrule
  1c3s5z          &9 \\  \midrule
  10m\_vs\_11m    &10 \\ \bottomrule
  \end{tabular}
  \label{tab:number-of-agents}
\end{table}

\begin{figure}[h]
  \centering
  \includegraphics[width=0.7\linewidth]{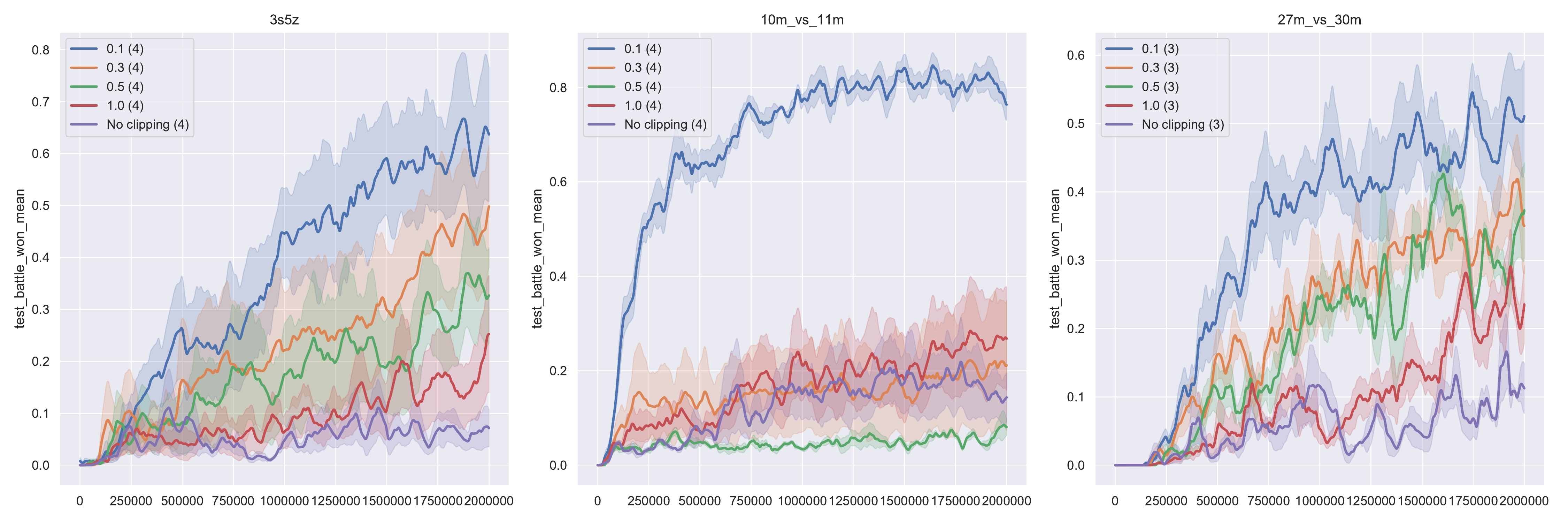}
  \includegraphics[width=0.7\linewidth]{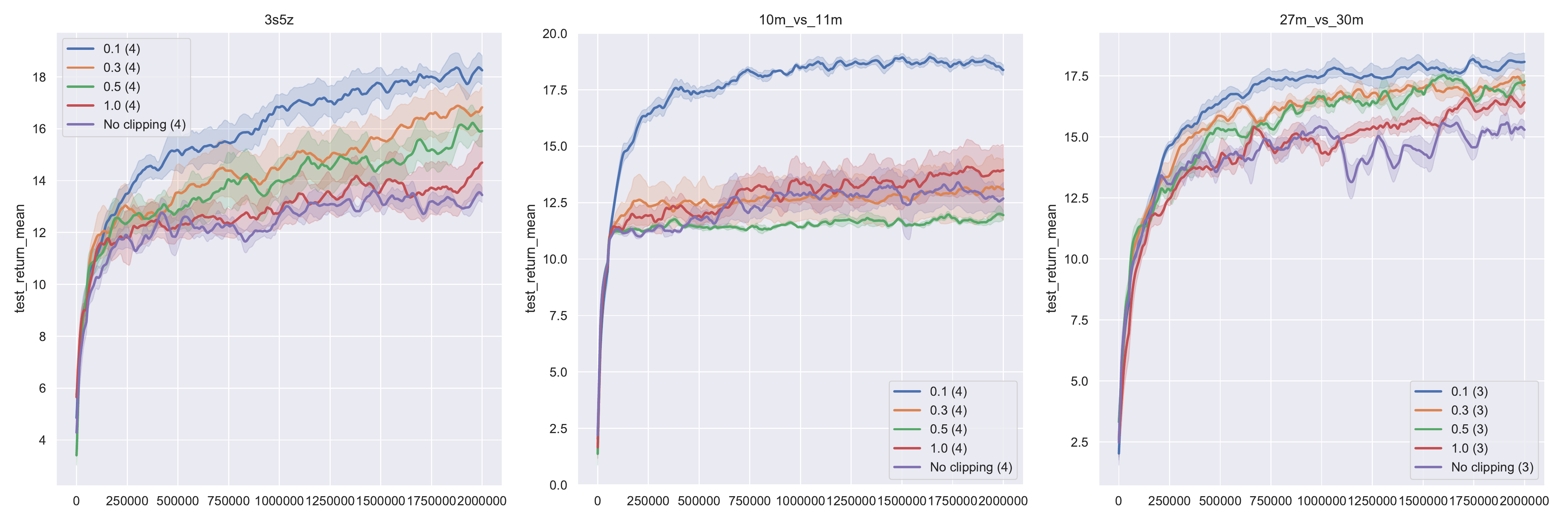}
  \includegraphics[width=0.7\linewidth]{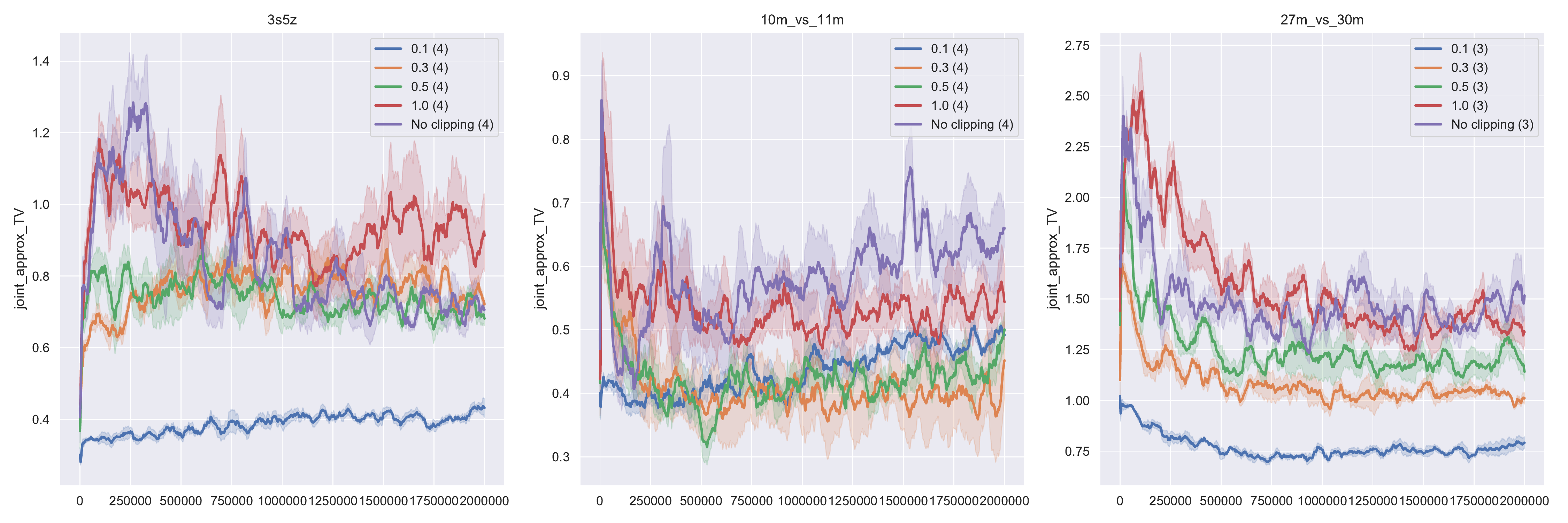}
  \caption{Empirical test battle win mean (first row), test returns (second row) and trust region estimates (third row) 
  of MAPPO on maps with varied difficult and numbers of agents}
  \label{fig:mappo-performance}
\end{figure}

Empirical test battle win mean, test returns and trust region estimates of MAPPO 
on maps with varied difficult and numbers of agents are presented in Figure~\ref{fig:mappo-performance}. 

\begin{figure}[h]
  \centering
  \includegraphics[width=1.0\linewidth]{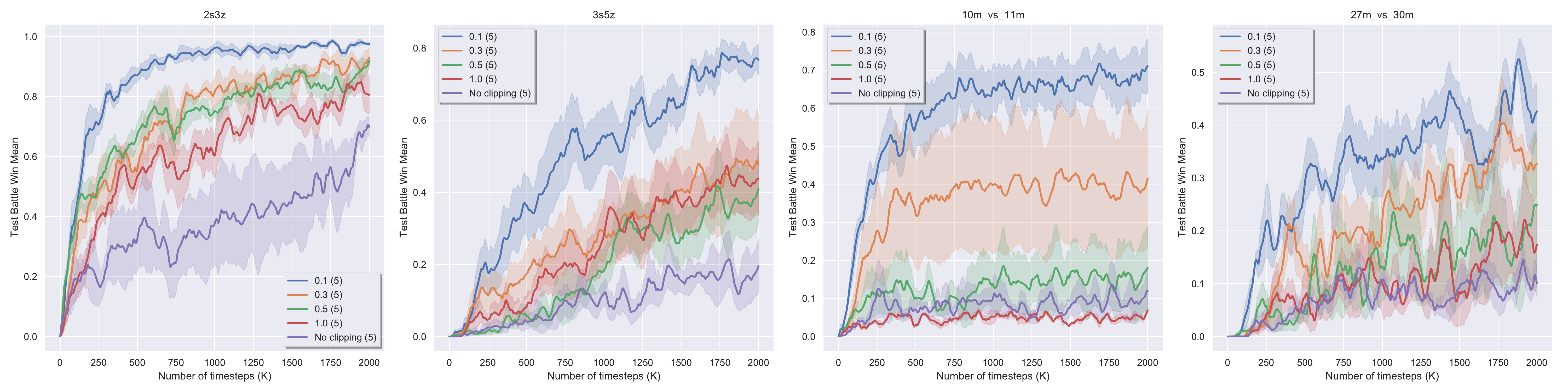}
  \caption{Test battle win mean of IPPO on maps with varied difficulty and numbers of agents}
  \label{fig:ippo-test-battle-win-mean}
\end{figure}

\begin{figure}
  \centering
  \includegraphics[width=0.30\linewidth]{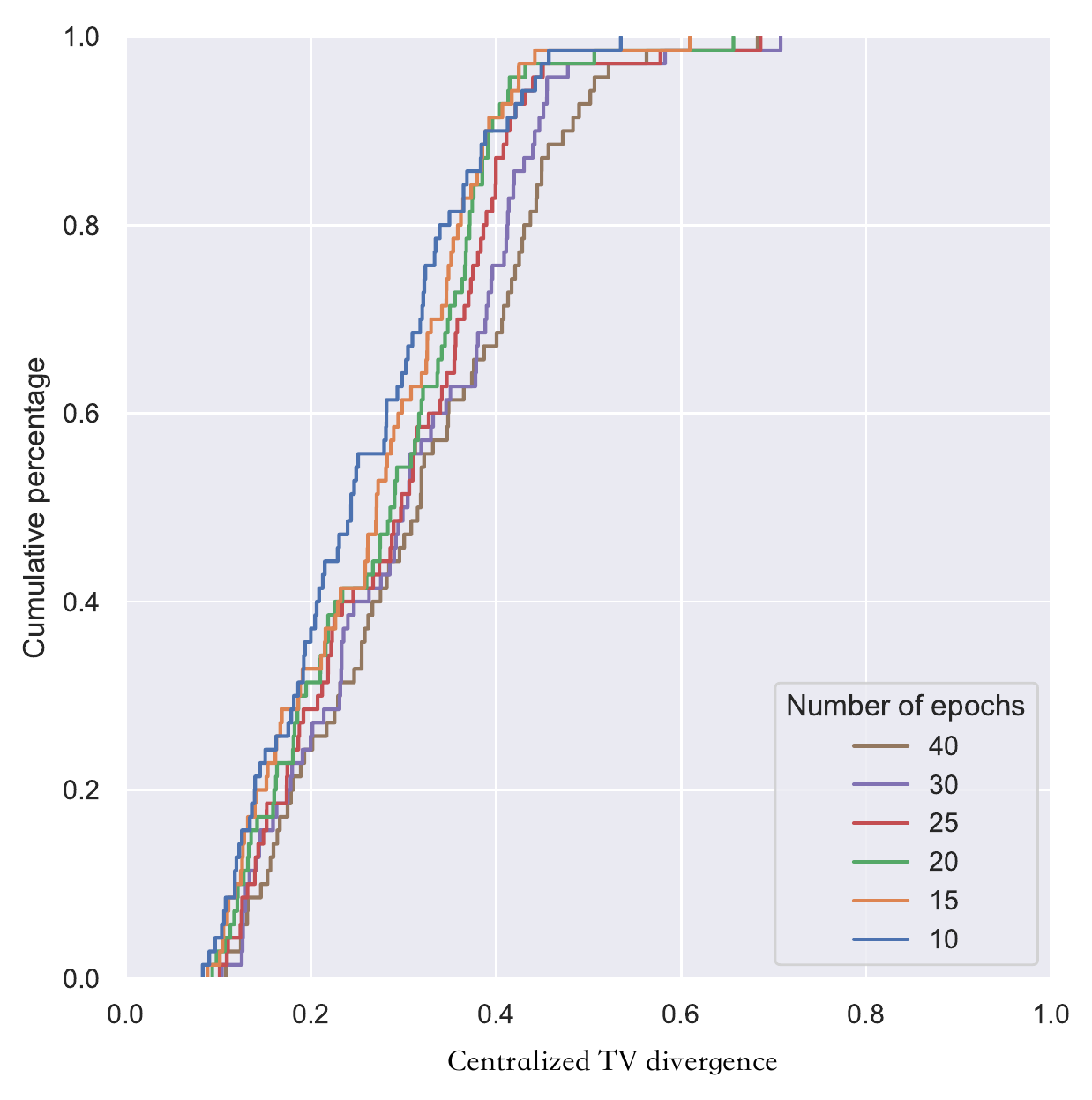}
  \caption{Cumulative percentage of centralized trust region with optimization epochs (clipping at $0.1$)}
  \label{fig:joint_trust_region_epoch}
\end{figure}

\subsection{Ablations on small clipping values}\label{app:small-clipping-range}
\begin{figure}
  \centering
  \includegraphics[width=0.7\linewidth]{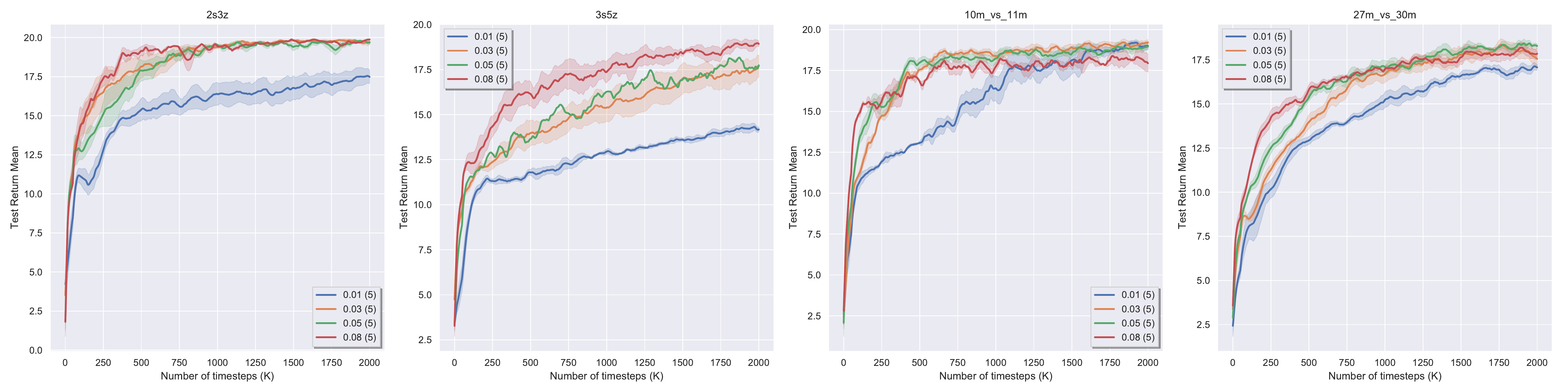}
  \includegraphics[width=0.7\linewidth]{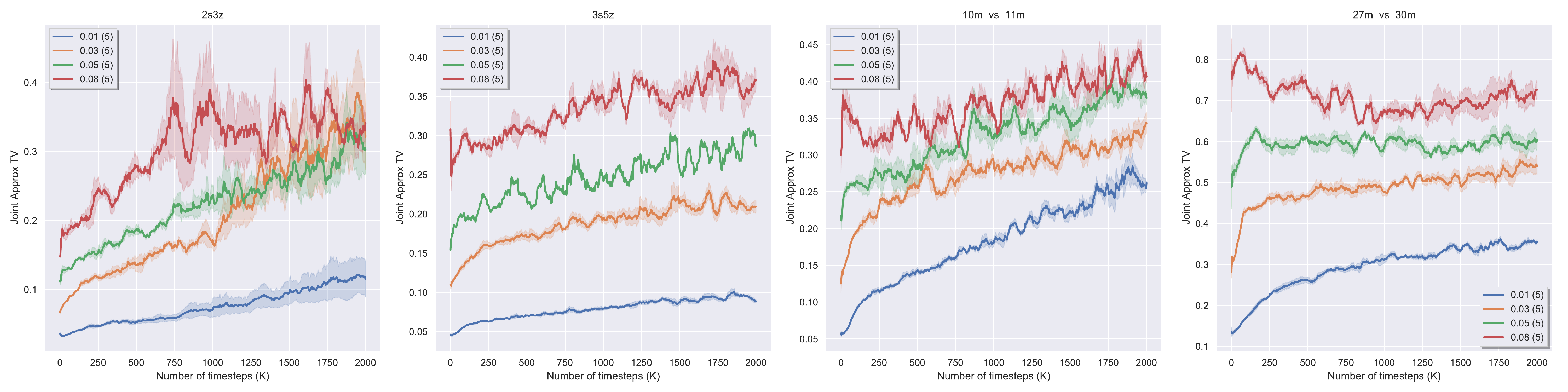}
  \includegraphics[width=0.7\linewidth]{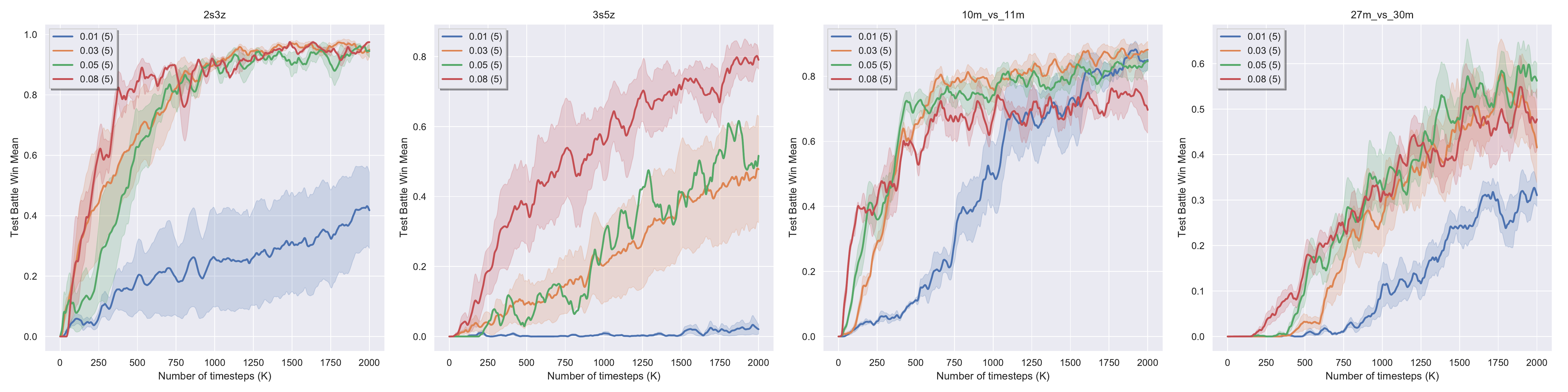}
  \caption{Empirical returns, trust region estimates and test battle win rate for small values of independent ratio clipping.}
  \label{fig:small-clipping-values}
\end{figure}

We also present the ablation results for small clipping values, i.e., $<0.1$, in Figure~\ref{fig:small-clipping-values}. 
It is true that a small clipping value results in a small trust region, 
and thus small clipping values, e.g., $0.08$, $0.05$ and $0.03$, 
would be preferred for maps with a large number of agents, 
e.g., maps 10m\_vs\_11m (10 agents) and 27m\_vs\_30m (27 agents).
However, when the clip value is too small, e.g., $\epsilon=0.01$ in maps with 5 and 8 agents,
the resultant trust region is also small and 
the update step in each iteration can thus be too small to improve the policy. 
Thus, one would need to trade off between the trust region constraint, 
to ensure monotonic improvement, 
and the policy update step, 
to ensure a sufficient parameter update at each iteration.


\end{document}